\documentclass{article}

\usepackage{fullpage}
\usepackage[small]{caption}
\usepackage{fancyhdr}
\usepackage{mathpazo}
\usepackage{natbib}
\usepackage{titling}

\usepackage[utf8]{inputenc} 
\usepackage[T1]{fontenc}    
\usepackage{hyperref}       
\usepackage{url}            
\usepackage{booktabs}       
\usepackage{amsfonts}       
\usepackage{nicefrac}       
\usepackage{microtype}      
\usepackage{xcolor}         
\usepackage{graphicx}       
\usepackage{multirow}       
\usepackage{tabularx}       
\usepackage{float}          
\usepackage{listings}       
\usepackage{comment}
\title{From Weights to Words: Expressing and Editing Preference Model Inferences in Natural Language}

\author{%
Zachary Wojtowicz\thanks{Preprint. Correspondence to \texttt{zachwoj@mit.edu}.} ~~~~ Ayush Nayak ~~~~ Jacob Andreas \\
MIT
}
\date{}

\usepackage{amssymb,amsmath,amsfonts,amsthm,algorithm,algpseudocode}

\usepackage{tikz}
\usetikzlibrary{positioning, arrows.meta, fit, calc}

\DeclareMathOperator*{\argmin}{arg\,min}

\definecolor{agentblue}{RGB}{78,121,167}   
\definecolor{agentred}{RGB}{225,87,89}     
\definecolor{agentgreen}{RGB}{89,161,79}   
\begin{document}

\maketitle

\begin{abstract}
  The growing use of statistical learning algorithms to infer human preferences from high-dimensional choice data runs up against a fundamental challenge: 
  choice alternatives typically differ in many ways simultaneously, so
  it is generally unclear which factors actually drove an observed decision and should be credited as preferences.
  Compounding this problem, the opacity of a preference model's inferences leaves human operators unable to inspect, contest, or correct them when they err.
  We introduce \emph{weights to words}, a method that takes a dataset of choice problems as input and automatically discovers a collection of domain-relevant preference dimensions, each described in natural language and paired with a vector in the model's representational space. 
  These dimensions address both under-determination and opacity: they can be applied to concentrate attribution on a small set of meaningful factors, and they can externalize the model's inferences in natural language so that users can inspect and edit them in real time.
  We first qualitatively illustrate the method's versatility on four diverse domains: moral dilemmas, movies, wines, and free-form LLM responses.
  We then report two pre-registered human-subjects experiments, on moral dilemmas ($N=450$) and movie selection ($N=449$), that demonstrate its benefits for learning preference models: (1) regularizing a preference model toward the learned basis increases prediction accuracy on held-out choices, and (2) incorporating participants' structured edits further improves accuracy. In head-to-head comparisons, participants prefer the method's inferred preference profiles and endorse its predictions as more accurate.
\end{abstract}

\section{Introduction}
\label{sec:intro}

The predominant approach to aligning frontier AI systems (exemplified by RLHF; \citealp{christiano2017deep})
trains large statistical models to predict human choice data, typically under the assumption that preferences can be represented by a linear utility function in the model's high-dimensional feature space.
Despite its many advantages, this strategy faces a fundamental problem:
inferring high-dimensional preferences from observed choices is often poorly
specified because alternatives differ along many---potentially thousands of---feature dimensions at once.
When someone chooses between two options that differ in many ways, which of those features was actually responsible for the choice?

This is one facet of a broader challenge that \citet{kleinberg2024inversion} call the ``inversion problem'': inferring latent mental states from observed behavior requires separating intentions from implementation noise, behavioral biases, habits, impulses, constraints, contextual influences, and innumerable other factors that also enter the choice process. Humans perform inversion so effortlessly that we do not even notice how integral it is to social reasoning. 
Strong inductive biases that encode the relevant psychology of decision making help separate someone's preferences from their choices by distinguishing the factors that actually drove their choice from the innumerable other ``insignificant'' differences between alternatives. We know intuitively, for example, that a student who chose to attend Brown University over Penn State likely preferred its small class sizes and liberal arts focus---not the fact that it comes earlier in the alphabet. However, the standard Bradley-Terry
\citep{bradley1952rank} preference model used to train LLMs and other
high-dimensional statistical models treats all such differences as equally
plausible contributors when they are represented in a common feature space, and
it is not clear in general how to place a prior over the parameters of a choice
model defined over embeddings produced by a neural network.

This challenge is compounded by the opacity of the distributed preference representations generated by these methods, which are fundamentally illegible to users, undermining transparency, accountability, contestability, and safety in deployment. This opacity reflects a deeper representational mismatch: models ``speak'' parameter updates, whereas humans are more adept at interpreting and expressing preferences in natural language, a tool refined over millennia for articulating our likes, wants, and needs. Ironically, although large language models excel at parsing language and mapping it to internal representations, methods for translating natural-language expressions into latent preference representations (and back) remain underdeveloped.

RLHF and supervised fine-tuning are not only statistically inefficient but can
also produce generalizations that are not just wrong but overtly unsafe, as
demonstrated by recent work on emergent misalignment. For example, an LLM
fine-tuned to write insecure code infers that it should behave maliciously far
beyond coding, even asserting that humans ought to be enslaved by AI
\citep{betley2025emergent,betley2025weird}. 
The authors of the latter paper call these ``weird generalizations,'' and they appear counter-intuitive precisely because gradient updates in a model's weight space are not constrained by human priors about what counts as a ``reasonable'' inference to draw from choice data. 

However, as we show in this paper, language models do actually learn such priors during pretraining: when prompted to explain choices \emph{in text}, LLMs reliably articulate reasonable, domain-relevant explanations for an observed choice. We exploit this asymmetry by extracting an interpretable, domain-relevant basis from these explanations, then translating it back into a model's representational space so it can be applied as a structural regularizer on learning dynamics and an interpretability bridge.

\paragraph{Contributions.}

We introduce an automated pipeline that takes as input a corpus of choice problems and discovers a low-dimensional basis of preference dimensions that can both regularize learning and structure how feedback is elicited from people. Each dimension is defined by a direction in a preference embedding space and a corresponding natural-language label, which makes it directly interpretable. We first illustrate the method's versatility by applying it to four choice domains---moral dilemmas, movies, wines, and LLM responses---and show that it recovers a plausible basis for each without domain-specific tuning (Section~\ref{sec:method}). We then demonstrate how such bases can be applied to improve preference learning in two pre-registered human-subjects experiments on moral dilemmas ($N=450$) and movies ($N=449$). In both domains, learning regularized with the basis significantly outperforms random projections, and natural-language feedback can further improve predictive accuracy and consistently improves subjective endorsement (Section~\ref{sec:behavioral}).

\section{Related Work}
\label{sec:related}

The standard alignment pipeline learns a scalar reward from pairwise comparisons and fine-tunes a policy against it, either through reinforcement learning or closed-form objectives \citep{christiano2017deep, ouyang2022training, rafailov2023direct}. 
Because binary preferences are informationally sparse, some recent work ``densifies'' the signal via, e.g., error-type labels on sub-sentence spans \citep{wu2023fine}. 
We inherit the BTL likelihood from this literature, but address sparsity by eliciting feedback in natural language and constraining inference to a low-dimensional, interpretable subspace.

A growing literature finds structured, human-readable directions in model representations and operates on them directly \citep{belinkov2022probing, rimsky2024steering, casademunt2025steering, gandelsman2023interpreting} or labels individual model components in natural language \citep{hernandez2021natural, choi2024scalingautomaticneuron}. 
\citet{movva2025s} specifically apply these methods for ``bottom-up'' feature labeling in datasets already annotated with human judgments.
We build on this body of work (and on standard sentence encoders, e.g., \citealp{reimers2019sentence}) by using LLMs to discover preference-relevant directions. Here, we are both interested in ``top-down'' discovery of meaningful dimensions that are not already captured by automated feature discovery techniques, and in using discovered features to improve the efficiency of human annotation itself.

A parallel line of work argues that a single universal reward function fails to capture the diversity of human preferences, proposing distributional, game-theoretic, or axiomatic alternatives \citep{siththaranjan2023distributional, halpern2025pairwise, munos2024nash, ge2024axioms, ge2024learning} and assembling large-scale pluralistic datasets \citep{kirk2024prism}. 
Our contribution is complementary, discovering a shared basis in which individualized preference models can be learned more efficiently.
Finally, recent work has explored post-training from naturalistic user interactions \citep{park2024rlhf, jin2025era, don2024naturally, liu2025user, stephan2024rlvf} or free-form conversations about user preferences \citep{li2023eliciting}; our work provides a structured representation of the preference function itself that could be combined with feedback signals derived using these methods.

\begin{figure}[t]
  \centering
  \includegraphics[width=\columnwidth,trim=0.1in 7.7in 2.1in 2.2in]{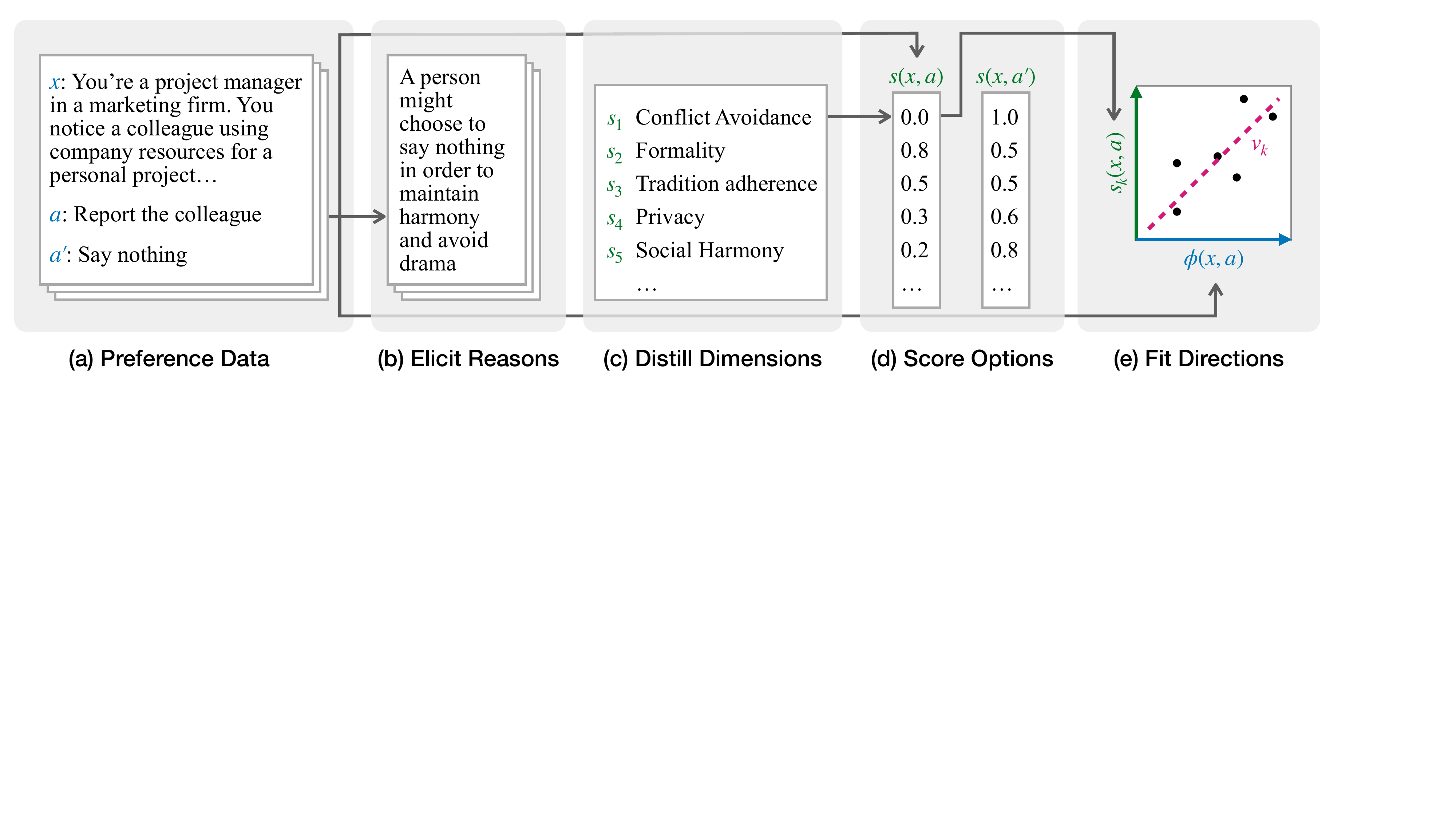}
  
  \caption{The discovery pipeline applied to moral dilemmas. (a) For each choice problem $(x, a, a')$, (b) an LLM produces free-text reasons to choose each side $a$ / $a'$. (c) LLMs condense these reasons into $K$ interpretable dimensions associated with names, summaries, and low and high poles (see Table~\ref{tab:dim-cards} for full examples). (d) Each choice is annotated with a scalar score $s_k(x, a)$ along each dimension $k$ by fitting a \emph{per-dimension} BTL model to pairwise LLM judgments. (e)
  Given high-dimensional choice representations (e.g., LLM embeddings),
  we find the representation subspace encoding each dimension by fitting a linear map from representations $\phi(x, a)$ to scores $s_k(x, a)$. These (choice dimension, representation subspace) pairs may be used directly for data analysis, to guide learning of preference models over $\phi$, or to elicit more informative preference annotations from humans.
  }
  \label{fig:method-overview}
  \end{figure}

\section{Method}
\label{sec:method}




\subsection{Preliminaries}
\label{sec:prelim}

We begin with the standard Bradley--Terry--Luce (BTL) stochastic preference model \citep{bradley1952rank} that underpins modern RLHF and related methods \citep{rafailov2023direct}. Let $\mathcal{X}$ denote a space of \emph{contexts} (e.g., descriptions of moral dilemmas) and $\mathcal{A}$ a space of \emph{actions} (e.g., things one could do or say in response). A \emph{choice problem} is a triple $(x, a, a') \in \mathcal{X} \times \mathcal{A} \times \mathcal{A}$. We assume a fixed feature map $\phi : \mathcal{X} \times \mathcal{A} \to \mathbb{R}^d$ given by a frozen language encoder. Define the difference vector
\begin{equation}
  \delta(x, a, a') \;=\; \phi(x, a) - \phi(x, a') \;\in\; \mathbb{R}^d.
\end{equation}
A linear utility $u_\theta(x, a) = \theta^\top \phi(x, a)$ with $\theta \in \mathbb{R}^d$ induces the BTL likelihood: an annotator's binary label $y \in \{0, 1\}$, with $y=1$ indicating $a \succ a'$, is distributed as
\begin{equation}
  y \mid x, a, a' \;\sim\; \mathrm{Bernoulli}\!\left(\sigma\bigl(\theta^\top \delta(x, a, a')\bigr)\right),
  \label{eq:btl-likelihood}
\end{equation}
where $\sigma(t) = (1 + e^{-t})^{-1}$ is the logistic function.\footnote{The choice of feature map is discussed in Section~\ref{sec:setup}.} 
Given a finite collection of choice problems and labels, our goal in this paper
is to learn a $\theta$ that predicts human judgments as accurately as possible
on new choices.

\begin{table}[t]
  \centering
  \small
  \begin{tabularx}{\linewidth}{@{}>{\raggedright\arraybackslash}p{2.4cm} >{\raggedright\arraybackslash}X >{\raggedright\arraybackslash}X@{}}
    \toprule
                            & \textbf{Low pole} & \textbf{High pole} \\
    \midrule
    \multicolumn{3}{@{}l}{\textbf{Privacy}: \emph{(moral dilemmas)}} \\[2pt]
    Pole label              & ``Public'' & ``Private'' \\
    Description             & ``Actions that involve public exposure or social visibility.'' & ``Actions that preserve personal boundaries and avoid public scrutiny.'' \\
    Typical person          & ``Someone who is comfortable with public attention.'' & ``Someone who values personal space and discretion.'' \\
    Example action          & ``Sharing personal experiences publicly'' & ``Keeping personal matters private'' \\
    \midrule
    \multicolumn{3}{@{}l}{\textbf{Emotional Depth}: \emph{(movies)}} \\[2pt]
    Pole label              & ``Emotionally Neutral'' & ``Highly Emotional'' \\
    Description             & ``Minimal character development or emotional engagement.'' & ``Rich character development and emotional resonance.'' \\
    Typical viewer          & ``A viewer who prioritizes plot over character arcs.'' & ``A viewer who seeks transformative storytelling.'' \\
    Example film            & \emph{Meet the Spartans} (2008) & \emph{The Color Purple} (1985) \\
    \midrule
    \multicolumn{3}{@{}l}{\textbf{Floral Aroma Intensity}: \emph{(wines)}} \\[2pt]
    Pole label              & ``Non-floral'' & ``Floral'' \\
    Description             & ``No detectable floral notes; more fruit or earthy character.'' & ``Prominent aromas of rose, lavender, or jasmine.'' \\
    Typical drinker         & ``A drinker who prefers bold or savory wines.'' & ``A drinker who enjoys aromatic whites or ros\'es.'' \\
    Example wine            & ``Barolo'' & ``Riesling (Alsace) or Gew\"urztraminer'' \\
    \bottomrule
  \end{tabularx}
  \vspace{2pt}\\
  \caption{Three example LLM-discovered preference dimensions, their corresponding pole descriptions, and an LLM-generated ``typical user'' with this preference (all generated by the method). Full per-domain dimension lists are in Appendix~\ref{app:dimensions}.}
  \label{tab:dim-cards}
\end{table}

\subsection{Automatic Discovery of Domain-Relevant Semantic Features}
\label{sec:pipeline}


We posit that, in many everyday choice domains, preferences are predominantly
driven by a relatively small subset of factors that span a low-dimensional
subspace of the ambient feature space, which we call the \emph{domain-relevant
subspace}. Given a corpus of $N$ choice descriptions for a domain (and their
encoder embeddings), our method automatically discovers a basis $V = [v_1,
\ldots, v_K] \in \mathbb{R}^{d \times K}$ of $K \ll d$ named directions that
define a human-legible coordinate system within a subspace of the ambient
feature space $\mathbb{R}^d$.

The pipeline runs in five stages, sketched in Figure~\ref{fig:method-overview}.

\begin{enumerate}

  \item \textbf{Stratified pair sampling.} We draw $S$ option pairs from the corpus, stratified into bands of cosine distance between embeddings $\phi(x, a)$. Stratification ensures the elicited reasons span the full range of plausible disagreements.

  \item \textbf{Reason elicitation.} For each pair $(a, a')$, we prompt an LLM to produce a fixed number of free-text reasons to choose $a$ and the same number to choose $a'$, in the framing of the domain (e.g., ``which action is more ethical?''). The output is a pool of free-text reasons.
    Full prompt templates (for this and subsequent steps) may be found in Appendix~\ref{app:dim-discovery}.

  \item \textbf{Dimension condensation.} 
  The reason pool is then distilled into a smaller set of canonical themes by LLM batched clustering.
  The top themes are passed to an LLM that emits $K$ dimensions, each with a
    name, and a description of users and choices aligned with low and high poles of this dimension. 
    (see Table~\ref{tab:dim-cards} for examples). 

  \item \textbf{Option scoring.} For each dimension $k = 1, \dots, K$ we sample a set $\mathcal{J}_k$ of option pairs 
    and fit a \emph{dimension-specific} BTL model to recover scalar scores $s_k(a_n) \in \mathbb{R}$ as the solution to
  \begin{equation}
    \argmin_{s_k} \sum_{(i,j) \in \mathcal{J}_k} -\log\sigma\!\big(
    (2y_{ij}^{(k)}-1) \,(s_k(a_i) - s_k(a_j))\big),
    \label{eq:bt-scoring}
  \end{equation}
  where $a_i$ and $a_j$ are the two options in pair $(i,j)$ and $y_{ij}^{(k)}
    \in \{0,\tfrac12,1\}$ is the LLM judge's verdict on whether $a_i$ scores higher than $a_j$ on dimension $k$, where the intermediate value $\tfrac12$ encodes a ``negligible'' verdict (the two options are indistinguishable on dimension $k$) and enters \eqref{eq:bt-scoring} as a tie.\footnote{A simpler variant (in which the LLM emits a discrete value per option) is described in Appendix~\ref{app:dim-discovery}. We use BTL judging as the primary method because we consistently found it to yield better ratings.}

  \item \textbf{Direction fitting in encoder space.} For each dimension $k$, we fit a direction $v_k \in \mathbb{R}^d$ that maps encoder embeddings $\phi$ to dimension scores $s$ by ridge regression:
  \begin{equation}
    v_k \;=\; \argmin_{v \in \mathbb{R}^d} \;\sum_{x,a} \bigl(s_k(x, a) - v^\top \phi(x, a)\bigr)^{\!2} \;+\; \alpha_k \|v\|^2,
    \label{eq:ridge}
  \end{equation}
  for some regularization parameter $\alpha_k$. Finally, we construct the preference basis $V \in \mathbb{R}^{d \times K}$ of discovered dimensions by $\ell_2$-normalizing and stacking the $v_k$.
\end{enumerate}

This pipeline harnesses three distinct LLM abilities: first, the \emph{reason elicitation} stage relies on the model's capacity to identify and articulate the relevant explanations for a choice in a given domain; second, the \emph{dimension condensation} stage relies on its capacity to abstract across those explanations, grouping semantically related ones and naming the underlying preference dimension they share; and third, the \emph{option scoring} stage requires the model to interpret a stated preference dimension and apply it consistently to choices, much as a human would. This last ability is especially important: it is what lets the method align each dimension with a direction in the encoder's embedding space that carries roughly the same meaning a person would assign it.



\subsection{Inference: Regularized BTL with a Feedback Prior}
\label{sec:inference}

We now describe how a participant's binary choices and (optional) structured feedback are elicited and combined to estimate their preference vector $\theta \in \mathbb{R}^K$ in the discovered subspace.
We begin by constructing a standard binary preference elicitation task (e.g., that used for RLHF). On each trial $t \in \{1, \ldots, T\}$ the participant chooses between two options $(a_t, a'_t)$ in context $x_t$. The response is recorded as a label $y_t \in \{0,1\}$. 

Our approach may then be used to guide preference learning in two different
ways:

\paragraph{Using the domain-relevant subspace as an inductive bias.} To confine
learning to the domain-relevant subspace, we project each trial's difference
vector $\delta_t$ onto the basis, $u_t = V^\top \delta_t \in \mathbb{R}^K$ and
stack these vectors into a design matrix $U \in \mathbb{R}^{T \times K}$ with
entries $u_{t,k}$. The signed entry $(2 y_t - 1)\,u_{t,k}$ captures what the
choice reveals about dimension $k$. We can then fit a preference model directly
in the space of projected representations $u$, ignoring components of
representations that did not correspond to a discovered preference dimension. In
experiments we refer to this method as \textbf{domain projection}.

\paragraph{Expressing model inferences in natural language.} 

The discovered dimensions can play an additional role in the preference learning interaction itself. 
\emph{After} each binary choice selection, we show annotators a set of provisional natural language statements expressing inferences that the model would draw from that observed choice, and ask them to affirm or edit each of these inferences.

Concretely, for each dimension $k$, we pre-compute feature values $u_{\cdot,k}$
for all options and group them into a set of discrete bins. We assign
each bin a natural language description reflecting its sign and magnitude
(\emph{prefer to skip}, \emph{aren't into}, \emph{indifferent}, \emph{like}, \emph{love}).\footnote{In our experiments the categories are symmetric quintile bins with midpoint representatives; see Appendix~\ref{app:rescaling}.}
On each trial $t$, we rank dimensions by the magnitude of their \emph{per-dimension–normalized} projection $|u_{t,k}|/m_k$, where $m_k = \max_{x,a}|u_{\cdot,k}|$ rescales each dimension onto a common range, and show the user the $M$ dimensions with the largest such values along with the description of each one's associated bin.

The participant responds to the displayed inferences in one of two modes.

In the
first mode, the participant affirms or removes each inference, producing an
additional feedback variable $z_{t,k} \in \{0, 1\}$.
See Fig.~\ref{fig:qt-affirm} for an example of this interface.
At the end of data collection, we construct a \emph{feedback-derived prior}:
\begin{equation}
\bar{\theta}_k = \frac{1}{T_k} \sum_t z_{t,k}\, \bar{u}_{t,k}
\end{equation}
where $\bar{u}_{t,k}$ is the representative value (the bin midpoint) of the bin to which $u_{t,k}$ is assigned, and $T_k$ is the number of trials on which dimension $k$ was surfaced to the participant.\footnote{Each dimension is normalized by its own count of feedback-bearing trials $T_k$ rather than the total trial count $T$, and the resulting raw prior is then rescaled to match the magnitude of the choice-derived estimate before it enters Equation~\eqref{eq:objective}; see Appendix~\ref{app:rescaling} for the full construction (bin boundaries, midpoint representatives, and rescaling).}
The values of $\bar\theta_k$ are used to guide learning of a preference model as
described below.  In experiments, we refer to this procedure as
\textbf{affirm/remove}.

In the second mode, the participant can freely re-assign each inference to a
different category, directly producing a feedback variable $\bar{u}'_{t,k}$. We
then construct the feedback prior:
\begin{equation}
\bar{\theta}_k = \frac{1}{T_k} \sum_t \bar{u}'_{t,k}
\end{equation}
where $T_k$ is again the number of trials on which dimension $k$ was surfaced (same normalization and rescaling as above).
In experiments, we refer to this procedure as \textbf{category select}.

Finally, we estimate the parameters $\theta$ of the preference model by minimizing the sum of a BTL negative log-likelihood, an isotropic L2 penalty, and a Gaussian prior centered at $\bar\theta$:
  \begin{equation}
    \hat\theta \;=\; \arg\min_{\theta \in \mathbb{R}^K}\;-\!\sum_{t=1}^T \log\sigma\!\bigl((2y_t-1)\,u_t^\top \theta\bigr) \;+\; \tfrac{\lambda}{2}\|\theta\|^2 \;+\; \tfrac{\mu}{2}\|\theta - \bar\theta\|^2.
    \label{eq:objective}
  \end{equation}
Setting $\mu=0$ recovers the projection-only estimator (no feedback); setting $\bar\theta=0$ recovers a pure ridge fit on the domain-relevant subspace. 

Hyperparameter values and optimizer settings are discussed in Section~\ref{sec:setup} and Appendix~\ref{app:implementation}.

\subsection{Implementation and Fit Evaluation}\label{sec:setup}

Throughout, we implement $\phi$ as a frozen Qwen3-Embedding-8B sentence encoder ($d=4096$); the pipeline's LLM-driven stages share a single instruction-tuned Qwen3-32B. Pipeline defaults and prompt templates are in Appendix~\ref{app:dim-discovery}; code and configurations are released with the paper.

To illustrate the general performance of the method, we first applied the weights-to-words pipeline to four domains: \emph{moral dilemmas} \citep{chiu2024dailydilemmas}, \emph{movies}, \emph{wines}, and \emph{LLM responses}, using $K=10$ for the first two and $K=15$ for the last two. Table~\ref{tab:domains} reports the bases with the metrics from Section~\ref{sec:decomp}: on moral dilemmas, movies, and wines the bases capture roughly $14$--$17\%$ of total choice variance (and $6\%$ on the harder LLM-responses domain), recover $47$--$58\%$ of the rank-$K$ PCA optimum ($14\%$ on LLM responses), and yield high per-dimension independence ($0.54$--$0.90$).

The remainder of this paper presents two sets of experiments: an
\emph{intrinsic} evaluation of the quality of the discovered bases
(Sec.~\ref{sec:decomp}), and an
\emph{extrinsic} evaluation of their usefulness for learning preference models
(Sec.~\ref{sec:behavioral}).

\section{Evaluating Discovered Bases}

\label{sec:decomp}

We evaluate three desirable characteristics of a domain-specific preference basis $V$: whether it (1) captures a large fraction of the variance in the choice domain; (2) spreads that variance across distinct dimensions; and (3) yields dimensions that are roughly independent. We benchmark $V$ against rank-$K$ principal components analysis (PCA), which is the variance-maximizing linear subspace of the same rank and therefore an upper bound on what any rank-$K$ basis could achieve in pure variance terms. 

Recall that $\delta(x, a, a') = \phi(x, a) - \phi(x, a')$. We evaluate variance against a reference distribution over \emph{contrasts}: let $C = \mathbb{E}[\delta\,\delta^\top]$ be the covariance of difference vectors under a chosen pairing distribution over options. Throughout we take this distribution to be two options drawn independently from the domain corpus, so that $C = 2\Sigma$ with $\Sigma = \mathrm{Cov}[\phi(x, a)]$ the covariance of option embeddings. This makes the metrics below a property of the basis and the domain corpus as a whole, rather than of any particular experiment's trial-pairing scheme.\footnote{Equivalently, $C$ is the covariance of the contrast between two options drawn at random from the corpus. For $d \gg N$ we recover its eigenvalues via the Gram-matrix identity $\mathrm{eig}(C) = 2\,\mathrm{eig}(X_c X_c^\top / N)$, where $X_c \in \mathbb{R}^{N \times d}$ is the centered embedding matrix, avoiding direct construction of $C$.} Let $\hat C = V^\top C V \in \mathbb{R}^{K \times K}$ be its projection onto the discovered subspace; here and in the metrics below $V$ denotes a QR-orthonormalization of the discovered directions, which span the same $K$-dimensional subspace but are individually $\ell_2$-normalized rather than mutually orthogonal, so that $\mathrm{tr}(\hat C)$ measures the variance captured by the orthogonal projection onto their span. Let $\lambda_1 \geq \lambda_2 \geq \cdots \geq \lambda_d$ denote the eigenvalues of $C$. We may then measure:

\begin{enumerate}
  \item \emph{Coverage.} The fraction of total choice variance the basis explains,
\begin{equation}
  \mathrm{Cov}(V) \;=\; \frac{\mathrm{tr}(\hat C)}{\mathrm{tr}(C)}.
\end{equation}
PCA upper-bounds this at $\mathrm{Cov}^\star_K = \sum_{j=1}^K \lambda_j / \sum_{i=1}^d \lambda_i$.

  \item \emph{Cumulative variance ratio.} How much of the rank-$j$ PCA optimum the $j$ highest-variance basis directions recover,
  \begin{equation}
    r_j \;=\; \frac{\sum_{i=1}^j v_i^\top C\, v_i}{\sum_{i=1}^j \lambda_i} \;\in\; [0, 1], \qquad j = 1, \ldots, K.
  \end{equation}
  Here the basis directions are indexed in order of \emph{decreasing} projected variance $v_i^\top C\, v_i$, so that $r_j$ compares the $j$ highest-variance discovered directions against the top-$j$ principal components. The trajectory $r_1, \ldots, r_K$ tracks how the discovered basis approaches the PCA frontier; values close to $1$ indicate the basis recovers nearly the full rank-$j$ subspace.

  \item \emph{Per-dimension independence.}
  \begin{equation}
    \mathrm{Indep}(v_j) \;=\; \frac{1}{[\hat C^{-1}]_{jj}\,\hat C_{jj}} \;\in\; [0, 1],
  \end{equation}
  which equals $1$ when $v_j^\top \delta$ is uncorrelated with the other projected components and approaches $0$ when $v_j$ is redundant given the rest of $V$.

\end{enumerate}

\begin{table}[t]
  \centering
  \small
  \caption{Evaluation of the pipeline in four domains: moral dilemmas, movies, wines, and LLM responses. Cov.\ is the fraction of choice variance captured by the rank-$K$ basis; $r_K$ is the cumulative variance ratio against the rank-$K$ PCA optimum; Indep.\ is the mean per-dimension independence (\S\ref{sec:decomp}).}
  \label{tab:domains}
  \begin{tabularx}{\linewidth}{@{}l c >{\raggedright\arraybackslash}X c c c@{}}
    \toprule
    Domain & $K$ & Example dimensions & Cov. & $r_K$ & Indep. \\
    \midrule
    Moral dilemmas & 10 & moral integrity, social harmony, informality, tradition adherence, efficiency, privacy, $\ldots$ & 0.151 & 57.8\% & 0.876 \\
    Movies & 10 & emotional depth, action intensity, humor intensity, historical authenticity, suspense/atmosphere, sci-fi/fantasy worldbuilding, $\ldots$ & 0.146 & 46.8\% & 0.835 \\
    Wines & 15 & fruit intensity, body and structure, acidity, oak influence, sweetness, aromatic complexity, $\ldots$ & 0.172 & 53.0\% & 0.902 \\
    LLM responses & 15 & conciseness, structure, actionability, clarity, emotional resonance, depth, $\ldots$ & 0.062 & 14.4\% & 0.540 \\
    \bottomrule
  \end{tabularx}
\end{table}

Table~\ref{tab:dim-cards} shows an example dimension for three domains to
illustrate the qualitative content of a discovered basis (pole labels,
descriptions, and an example option); Table~\ref{tab:domains} reports
qualitative metrics across all four domains. See
Appendix~\ref{app:dim-discovery} for additional details. In general we find that
dimensions capture important qualitative features, exhibit a high degree of
independence, and explain a large fraction of the variance that would be
captured by a comparably-sized PCA subspace.

\begin{figure}[t]
  \centering
  \includegraphics[width=0.65\linewidth]{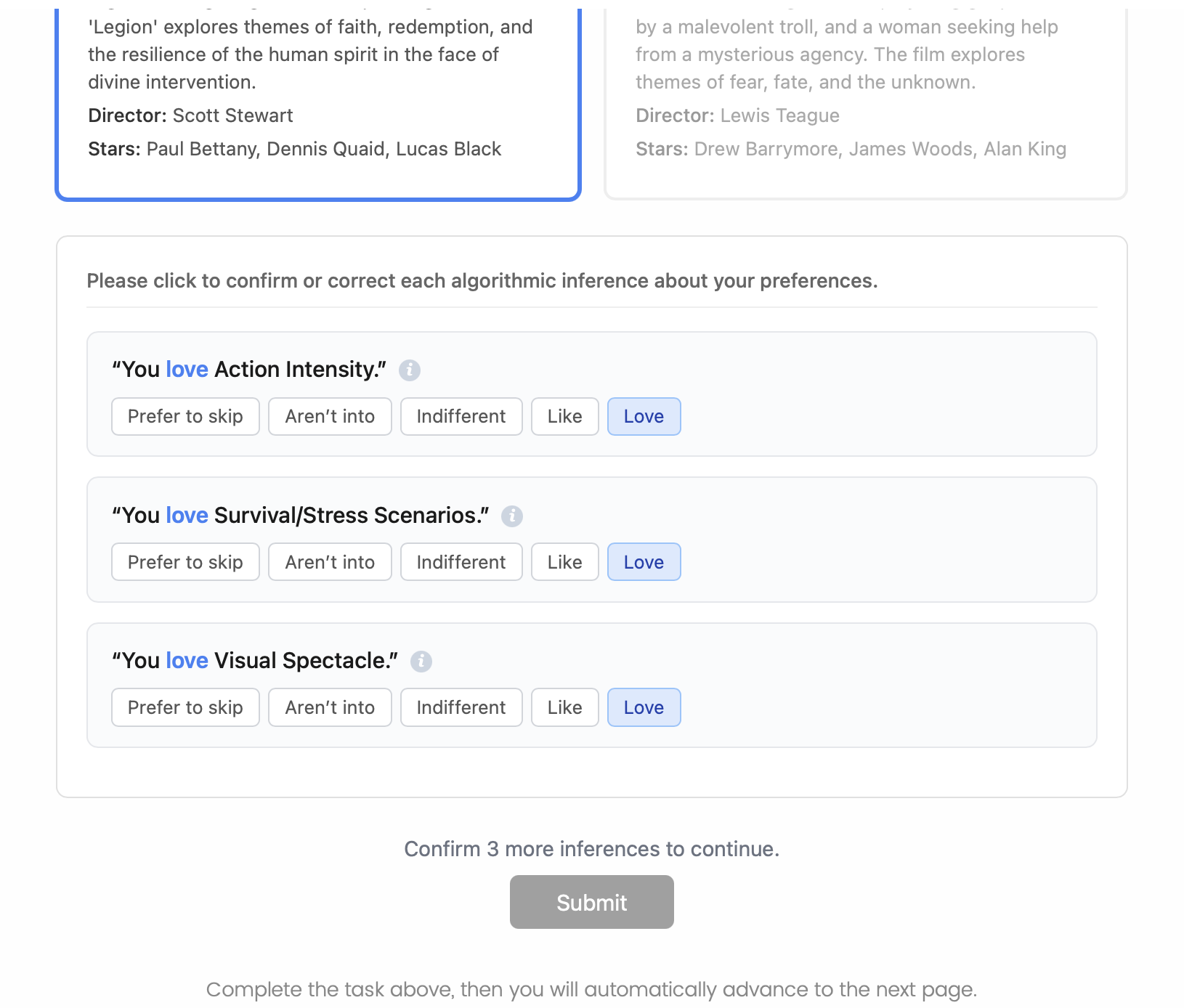}
  \caption{Feedback panel in the \textsc{inference\_categories} condition (movies). For each of the top-3 dimensions, the model pre-selects a level on a five-level scale ($\{$skip, not into, indifferent, like, love$\}$); the participant can change the pre-selected level before submitting.}
  \label{fig:qt-categories}
\end{figure}

\section{Behavioral Experiment}
\label{sec:behavioral}

Our main experiment evaluates the impact of using a weights-to-words basis for
learning in two behavioral experiments, one on moral dilemmas ($N=450$) and a
parallel study on movies ($N=449$). 
The experiment asks (i) whether the LLM-discovered semantic basis aids
preference inference relative to a same-rank random projection, and (ii) whether
augmenting binary choices with natural-language feedback further improves
recovery and subjective endorsement.

\subsection{Design}
\label{sec:behav-design}

The experiment was hosted on Qualtrics.\footnote{The full Qualtrics integration
script is included with the released code (Appendix~\ref{app:implementation}).}
Participants were recruited via Prolific. The experiment was pre-registered on
AsPredicted. For the moral-dilemmas study, $N=450$ participants completed the
full procedure and met the pre-registered inclusion
criteria.\footnote{\url{https://aspredicted.org/z3qu4z.pdf}} For the parallel
movies study, $N=449$ participants completed the procedure under the same
inclusion rules.\footnote{\url{https://aspredicted.org/n8m8yr.pdf}} Sessions
took a median of 14:11 (dilemmas) and 9:54 (movies). Participants were paid a
show-up fee of $\$3.75$ through Prolific, so final compensation corresponds to
roughly \$16/hr on dilemmas and \$23/hr on movies. The protocol was reviewed and
evaluated as an exempt behavioral intervention by our institutional IRB.

The corpora and bases are two of those described in Subsection~\ref{sec:setup}.
The first is a collection of moral dilemmas drawn from the dataset DailyDilemmas
\citep{chiu2024dailydilemmas}, deployed with the choice prompt: ``Which
action is more ethical or understandable?'' The second is a sample of 100 films
drawn from the MovieLens-32M dataset \citep{harper2015movielens}, deployed with
the prompt: ``Which movie would you rather watch right now?''\footnote{MovieLens supplies each film's title and genres; we additionally retrieve its director and top-billed cast from the TMDB API and generate a short (2--4 sentence) plot summary with an instruction-tuned LLM grounded on the title, genres, and TMDB overview. The encoder $\phi$ then embeds a short text card assembling these fields.} Both bases use $K=10$ LLM-discovered dimensions (e.g., moral integrity, social harmony, tradition adherence for dilemmas; emotional depth, action intensity, humor intensity for movies; full lists in Appendix~\ref{app:dimensions}). Candidate trial pairs were drawn stratified by cosine distance so participants saw a mix of close and contrasting pairs.

Each session ran in five phases: (i) \emph{consent}; (ii) $T_{\text{prac}}=5$
\emph{practice} trials with a single dimension shown per trial and immediate
correctness feedback, to familiarize participants with the dimensions and their
application; (iii) the \emph{main task} of $T=20$ binary-choice trials, with the
top-$M=3$ dimensions (ranked by the per-dimension--normalized projection $|u_{t,k}|/m_k$, as in Section~\ref{sec:inference}) surfaced beneath the options in conditions that elicited structured feedback.
After the main task, a preference model was fit to the participant's annotations
as described in Sec.~\ref{sec:inference}. The predictions from this preference model were presented to the participant in two evaluation tasks: (iv) a head-to-head \emph{summary
comparison}, in which participants were asked to judge each model's inferred preference profile, which included the top-5 (most positive) and bottom-5 (most negative) dimensions; and (v) two \emph{held-out prediction} trials---one per model---in which participants rated the accuracy of each model's predicted choice on a pair where the two models disagreed and one was confident.\footnote{Specifically, pairs are ranked by: unseen before seen; then whether the models predict \emph{opposite} options (directional disagreement); then the larger of the two models' utility margins. The top two pairs are rated, one under each model. Disagreement is thus prioritized but not strictly guaranteed, and the final sort selects for one model being confident.}

Participants were randomly assigned to one of three conditions.
In \textsc{choice\_only}, no dimensional information is shown; this condition
was used to evaluate the usefulness of the domain projection procedure. 
In \textsc{inference\_affirm}, the model proposes a natural-language inference for each of the top-3 dimensions and the participant affirms or removes each. In \textsc{inference\_categories}, the participant rates each of the top-3 dimensions on a five-level scale $\{\text{skip}, \text{not into}, \text{indifferent}, \text{like}, \text{love}\}$, with each level mapped to a per-dimension magnitude via the quintile-midpoint scheme of Appendix~\ref{app:rescaling}. The same three conditions ran in both domains.

Each condition then compares the prediction of two models, which we refer to as
the \textbf{augmented} and \textbf{baseline} models.
For \textsc{choice\_only}, the augmented model uses the discovered $V$
($\mu{=}0$), while the baseline uses a fixed random orthonormal $V_{\text{rand}}$ of equal rank, shared across participants ($\mu{=}0$). For the inference conditions, augmented adds the feedback prior ($\mu{>}0$); baseline is projection-only on the same $V$ ($\mu{=}0$).
Importantly, when reading results below, improvements over baseline for the
\textsc{choice\_only} condition show that domain projection improves
over ordinary preference learning; improvements over baseline for the
\textsc{inference} conditions reflect that these procedures improve \emph{over
domain projection}.

We pre-registered three within-subjects hypothesis tests:
\begin{enumerate}
  \item[\textbf{H1}] \emph{Predictive accuracy.} The (leave-one-out) accuracy of the
    augmented model is greater than the baseline (paired one-sided $t$-test).
  \item[\textbf{H2}] \emph{Summary preference.} Participants
    prefer the preference summary generated by the augmented model over the baseline (one-sample Wilcoxon, alternative ${>}0$).
  \item[\textbf{H3}] \emph{Prediction endorsement.}      The
    augmented model's prediction is preferred to the baseline's, on held-out pairs selected so the two models disagree (one-sample Wilcoxon on the paired rating difference, alternative ${>}0$).
\end{enumerate}

When evaluating H1, we select hyperparameters using participant-level $K{=}5$ cross-validation. 
H2 and H3 reflect subjects' judgments of the summaries and predictions rendered
live in the participant's browser during the task, at the per-domain parameters
the Qualtrics interface used at runtime. 
A parallel H1 analysis at these deployed parameters appears in
Appendix~\ref{app:deployed-results}, and details of the cross-validation
procedure may be found in Appendix~\ref{app:calibration}.  We apply a
Holm-Bonferroni correction when evaluating all hypotheses.

\subsection{Results}
\label{sec:behav-results}

\begin{figure}[t]
  \centering
  \includegraphics[width=0.98\linewidth]{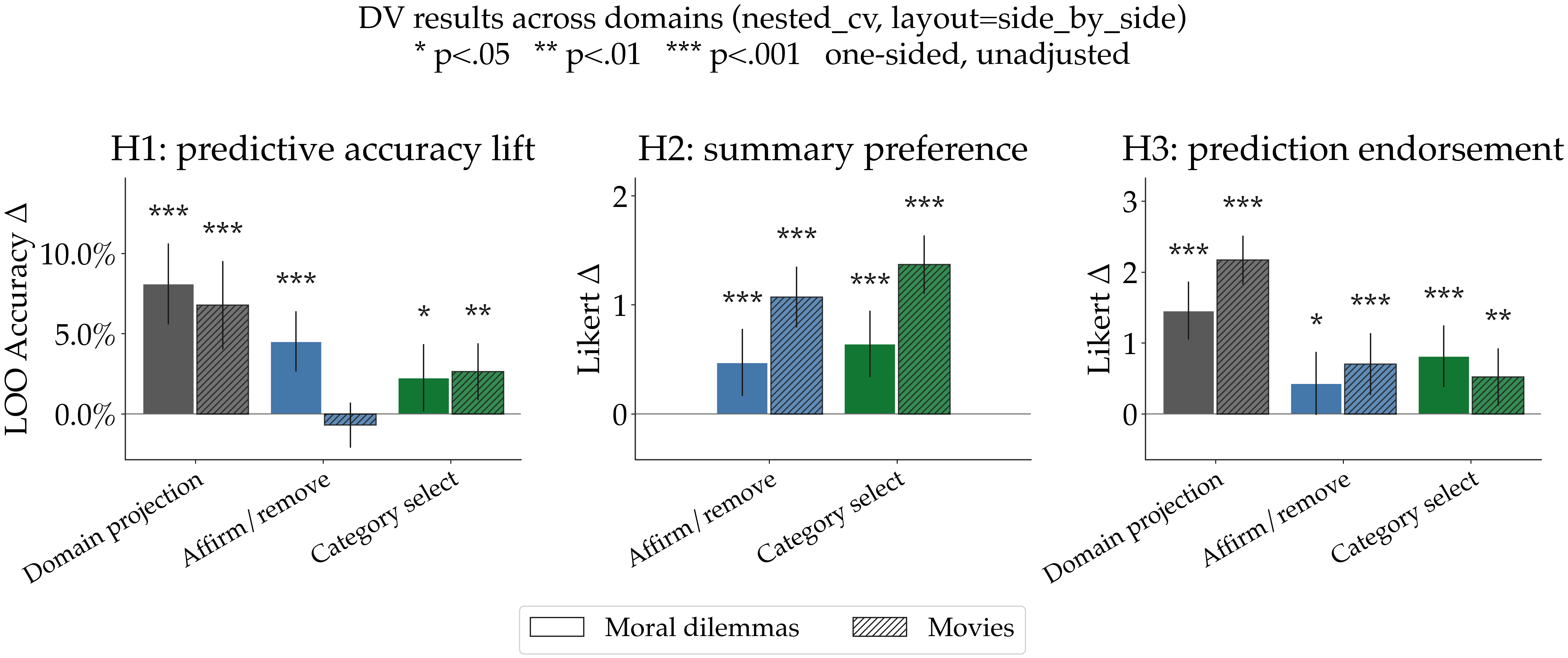}
  \caption{Behavioral results on moral dilemmas ($N=450$, solid bars) and movies ($N=449$, hatched bars). \textbf{H1} (LOO accuracy increase, augmented minus baseline, in percentage points) reports per-participant held-out lift under nested $K{=}5$ participant-level cross-validation: the inner search picks $(\text{multiplier scheme},\,\mu,\,\lambda)$ on outer-train participants (or $\lambda$ alone for \textsc{choice\_only}, whose contrast has no prior), and each outer-test participant contributes one $\Delta$acc scored at that chosen cell. Note that the ``Affirm / remove'' and ``Category select'' conditions represent changes relative to domain projection. \textbf{H2} (signed 6-point Likert summary preference, positive favors augmented; shown for the inference conditions only, as \textsc{choice\_only} is omitted from H2) and \textbf{H3} (paired prediction-rating difference) reflect participants' judgments of summaries and predictions that were rendered live at the experimentally deployed parameters and cannot be re-fit post-hoc. Bars are condition means with 95\% CI; per-condition $N$s and exact $p$-values appear in Table~\ref{tab:behav-main}. The parallel analysis with H1 also evaluated at the deployed parameters is reported in Appendix~\ref{app:deployed-results} (Figure~\ref{fig:behav-deployed}, Table~\ref{tab:behav-deployed}).}
  \label{fig:behav-main}
\end{figure}

\begin{table}[t]
  \centering
  \small
  \begin{tabular*}{\linewidth}{@{\extracolsep{\fill}}lrrrr@{}}
    \toprule
    Condition & $N$ & H1 ($\Delta$acc pp, $p_{\text{holm}}$) & H2 (mean, $p_{\text{holm}}$) & H3 (mean $\Delta$, $p_{\text{holm}}$) \\
    \midrule
    \multicolumn{5}{@{}l}{\emph{Moral dilemmas}} \\
    \textsc{choice\_only}            & 153 & $+8.1$, $<\!.0001$ & \multicolumn{1}{c}{---} & $+1.46$, $<\!.0001$ \\
    \textsc{inference\_affirm}       & 146 & $+4.5$, $<\!.0001$ & $+0.47$, $.001$          & $+0.43$, $.025$ \\
    \textsc{inference\_categories}   & 151 & $+2.3$, $.018$     & $+0.64$, $.0001$         & $+0.81$, $.0005$ \\
    \midrule
    \multicolumn{5}{@{}l}{\emph{Movies}} \\
    \textsc{choice\_only}            & 152 & $+6.8$, $<\!.0001$ & \multicolumn{1}{c}{---} & $+2.17$, $<\!.0001$ \\
    \textsc{inference\_affirm}       & 151 & $-0.7$, $.836$     & $+1.07$, $<\!.0001$      & $+0.70$, $.002$ \\
    \textsc{inference\_categories}   & 146 & $+2.6$, $.004$     & $+1.37$, $<\!.0001$      & $+0.52$, $.007$ \\
    \bottomrule
  \end{tabular*}
  \vspace{.25em}
  \caption{Pre-registered behavioral results on moral dilemmas ($N=450$) and movies ($N=449$). H1 reports per-participant LOO accuracy lift in percentage points (augmented minus baseline) pooled across outer folds of nested $K{=}5$ participant-level cross-validation (hyperparameters chosen on outer-train, scored on outer-test). H2 reports signed 6-point Likert summary preference of the summaries shown to participants during the experiment; H3 reports paired prediction-rating difference for the predictions shown. H2 and H3 reflect judgments of summaries / predictions rendered at the experimentally deployed parameters and cannot be re-fit post-hoc. All $p$-values one-sided; $p_{\text{holm}}$ adjusted within each hypothesis family. \textsc{choice\_only} is omitted from the H2 summary-preference analysis (shown as ---). The parallel table with H1 also at the deployed parameters is in Appendix~\ref{app:deployed-results} (Table~\ref{tab:behav-deployed}).}
  \label{tab:behav-main}
\end{table}

\begin{figure}[t]
  \centering
  \includegraphics[width=0.98\linewidth]{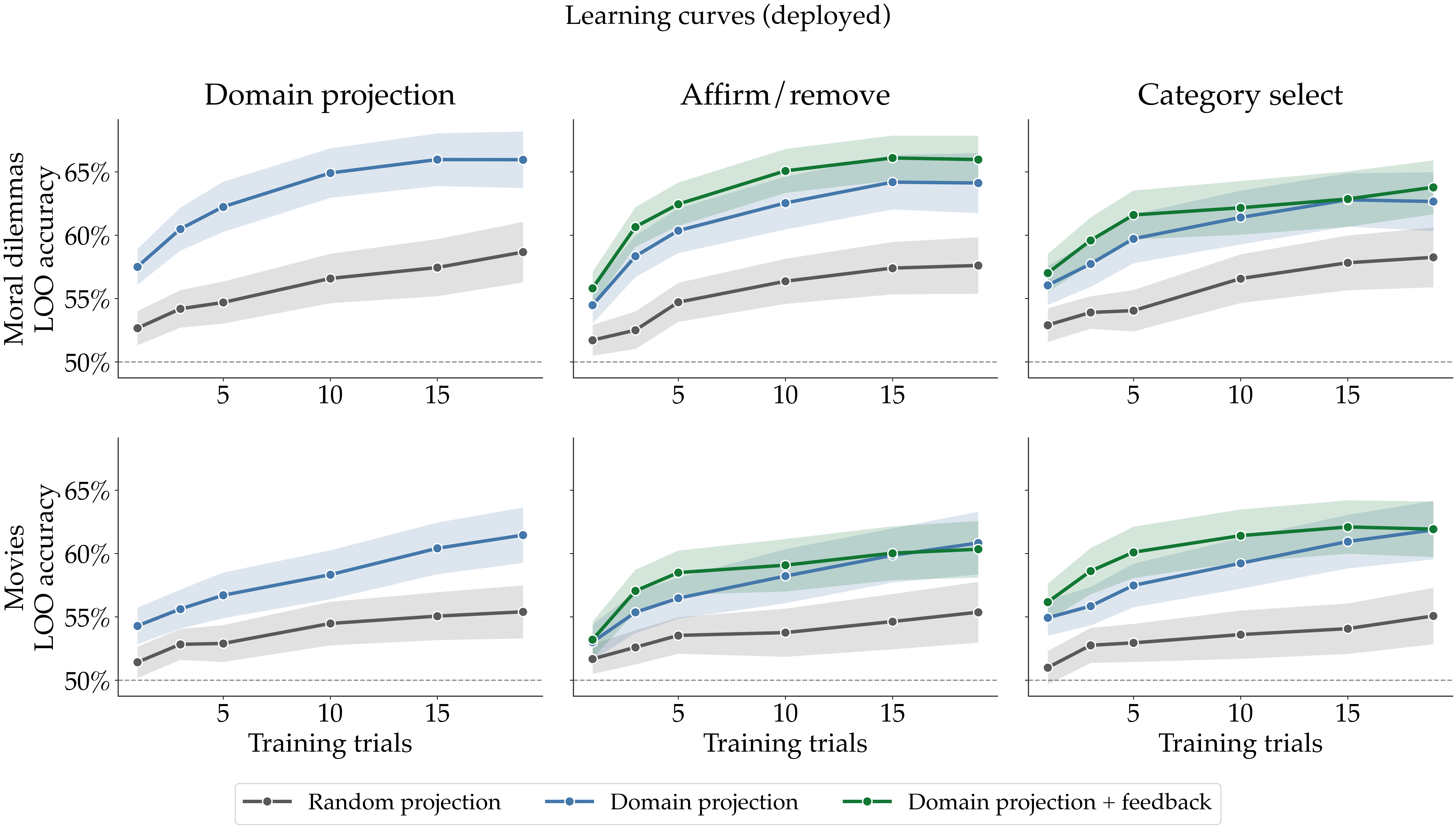}
  \caption{Learning curves on moral dilemmas (top row) and movies (bottom row). Per-condition LOO accuracy (\%) as a function of training-set size for random projection (gray), domain projection (blue), and domain projection $+$ feedback prior (green), with 95\% CI bands; dashed line marks chance. The feedback prior is most valuable when training data is sparse and converges with the projection-only model as more choices are observed.}
  \label{fig:learning-curves}
\end{figure}

Results are shown in Figure~\ref{fig:behav-main} and Table~\ref{tab:behav-main}. Four findings stand out.

\emph{First, the accuracy effect of projection is large and replicates across domains.} The LLM-discovered basis improves held-out predictive accuracy by $+8.1$ percentage points on dilemmas and $+6.8$ on movies over a same-rank random projection (\textsc{choice\_only} row, $p<.0001$ on both), and participants endorse its predictions over the baseline's on both domains (H3, $p<.0001$). The discovered basis therefore confers beneficial inductive bias beyond what dimensionality reduction alone provides.

\emph{Second, augmenting choices with natural-language feedback further improves
held-out accuracy on dilemmas.} On dilemmas the feedback prior adds a
substantial held-out accuracy lift over projection-only:
\textsc{inference\_affirm} reaches $+4.5$ \% ($p{<}.0001$, $d_z{=}+0.39$) and
\textsc{inference\_categories} reaches $+2.3$ \% ($p{=}.018$, $d_z{=}+0.17$);
the inner search in nearly every outer fold prefers a small prior strength ($\mu
\in [0.01,\,1.0]$) and a small $\lambda$ ($\lambda{=}0.001$), confirming the
calibration story that the deployed $\mu{=}2.0$ was over-shrinking. On movies,
the H1 picture diverges: \textsc{inference\_categories} reaches a modest but
reliable $+2.6$ \% ($p{=}.004$, $d_z{=}+0.24$), while \textsc{inference\_affirm}
delivers no H1 lift at all ($-0.7$ \%, $p{=}.84$, $d_z{=}-0.08$). Synthetic-participant simulations in Appendix~\ref{app:simulation} reproduce the same qualitative pattern under each domain's runtime parameters.

\emph{Third, the subjective measures (H2 and H3) are uniformly positive, and the
summary-preference effect (H2) is roughly twice as large on movies as on
dilemmas.} On every inference condition $\times$ domain cell, augmented
summaries and predictions are preferred over baselines (all $p{<}.05$, half
$<\!.001$). H2 effects on movies are about $2.2\times$ those on dilemmas (affirm
$+1.07$ vs.\ $+0.47$; categories $+1.37$ vs.\ $+0.64$); H3 patterns are mixed,
with no consistent cross-domain gap (affirm: movies $+0.70$ vs.\ dilemmas
$+0.43$; categories: movies $+0.52$ vs.\ dilemmas $+0.81$). The H2 size
differential is consistent with movies being the more articulable domain. 

\emph{Fourth, predictive accuracy (H1) and subjective preference (H2/H3) separate on movies.} \textsc{inference\_affirm} on movies delivers no H1 lift ($-0.7$ pp, $p{=}.84$) yet shows a large H2 effect ($+1.07$, $p{<}.0001$) and a significant H3 effect ($+0.70$, $p{=}.002$). Participants endorsed and preferred summaries that did not (per cross-validation) improve held-out accuracy at all over projection-only. 

\subsection{Sample-Efficiency Analysis}
\label{sec:behav-curves}

Our pre-registered hypothesis tests target a single accuracy number per participant at the full $T=20$ choices. To examine \emph{when} during a session each
estimator's gains emerge, we re-fit each model at every training-set size
$T'\in\{1,\ldots,T-1\}$ and compute LOO accuracy on the remaining trials.
Figure~\ref{fig:learning-curves} shows the resulting learning curves at the
experimentally deployed hyperparameters of each domain. 
Three patterns are consistent with the underdetermination story: the random-projection baseline plateaus throughout, the domain-projection model rises steadily as choices accumulate, and the projection-plus-feedback model starts substantially above projection-only at small $T'$ before the two converge as choice information accumulates. The feedback prior thus contributes most where binary choices alone provide the least signal and inductive bias is most valuable. Synthetic-participant simulations at the same hyperparameters (Appendix~\ref{app:simulation}, Figures~\ref{fig:sim-dilemmas-curves}--\ref{fig:sim-movies-curves}) reproduce the same monotonic ordering and the same convergence behavior at large $T'$.

\subsection{Comparison Against the PCA Subspace}
\begin{figure}[t]
  \centering
  \includegraphics[width=0.66\linewidth]{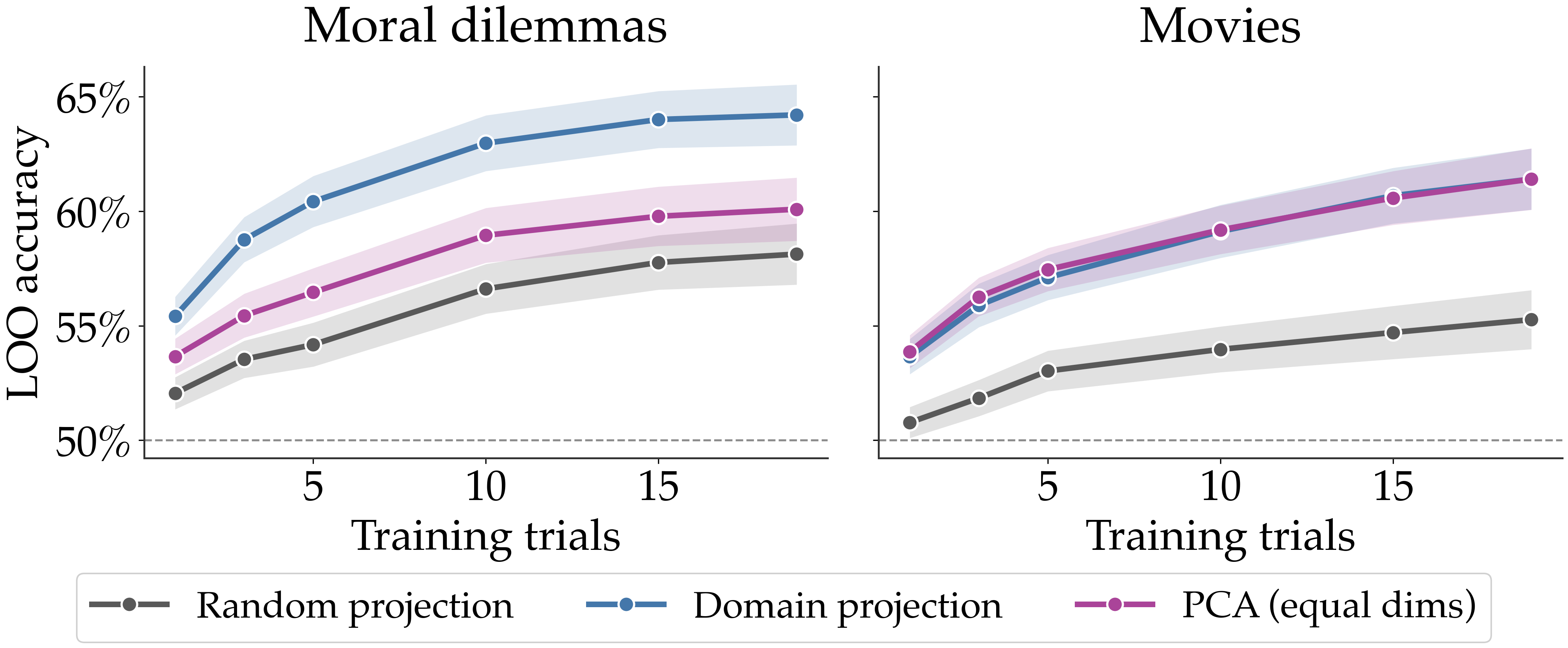}
  \caption{\textbf{The interpretable projection matches the variance-optimal PCA
  subspace on prediction.} Per-participant leave-one-out choice accuracy as a
  function of training-set size $T'$ on moral dilemmas ($N=450$) and movies
  ($N=449$), for three rank-$K$ estimators: a same-rank \emph{random} projection
  (gray), our LLM-discovered \emph{domain} projection (blue), and the rank-$K$
  \emph{PCA} subspace---the variance-optimal linear basis of equal rank
  (\S\ref{sec:decomp}, purple). Evaluation is identical to
  Figure~\ref{fig:learning-curves} (same leave-one-out fitter and train-size
  grid, $\mu{=}0$), here pooled across conditions; bands are 95\% CIs across
  participants and the dashed line marks chance. The domain projection leads
  on dilemmas and is on par with PCA on movies, while both dominate the random
  baseline.}
  \label{fig:estimator-comparison}
\end{figure}

Section~\ref{sec:decomp} benchmarks the discovered basis against rank-$K$ PCA
(i.e., the variance-optimal linear subspace of equal rank) as an upper bound on
\emph{variance} coverage. Here we ask the predictive analogue: does constraining
inference to the interpretable basis cost held-out accuracy relative to that
variance-optimal baseline? Figure~\ref{fig:estimator-comparison} compares the
domain projection, the rank-$K$ PCA subspace, and a same-rank random
projection under the \emph{same} leave-one-out evaluation as
Figure~\ref{fig:learning-curves} (same fitter and train-size grid, $\mu{=}0$),
here pooled across conditions. Here, domain projection matches or exceeds PCA.

\section{Discussion}
\label{sec:discussion}

In this paper, we introduce \emph{weights to words}, a method to identify and apply human-interpretable features of choice domains to improve inference. Section~\ref{sec:behavioral} shows what natural-language structure adds: on a named, low-dimensional basis, binary choices recover individual preferences from a handful of trials, and augmented summaries and predictions are reliably preferred over baselines on both domains.


One limitation is that the pipeline requires a pre-specified domain and corpus, and we did not study cross-domain transfer of $V$ in our analyses. LLM-discovered dimensions may inherit biases from the labeling LLM. On this point, we report diagnostics (Section~\ref{sec:decomp}, Appendix~\ref{app:dim-discovery}) but do not audit the bases against an external taxonomy. Another limitation is that pre-registered validation covers only two of our domains, with the others (wines and LLM responses) evaluated offline only. Finally, fitting calibration is domain-specific: the optimal multiplier scheme, prior strength $\mu$, and regularization $\lambda$ vary across our two online domains (Appendix~\ref{app:calibration}). The body figure handles this by selecting hyperparameters via nested participant-level cross-validation; a fresh domain therefore requires a small pilot to support the inner sweep, and the at-deployed-parameters analysis in Appendix~\ref{app:deployed-results} shows how much the conservative live-preview defaults under-state the method's predictive lift.

Three directions for future work stand out. The first is active elicitation, or selecting trials adaptively to minimize expected posterior entropy on $\theta$. The second is to explore deeper bases by applying the discovery pipeline to LLM hidden-state directions rather than encoder features, enabling basis-constrained fine-tuning at the activation level. The last would be to study hybrid feedback, such as combining binary choices, natural-language inferences, and slider ratings in one Bayesian fitter so participants use whichever modality is most articulable. All three slot into the projection-onto-$V$ framing without changes to the inference machinery.

The approach developed in this paper confronts a fundamental problem in aligning algorithms with human preferences: as more capable systems train on ever more complex choice data, the need for stronger inductive biases grows as well. Weights-to-words takes a first step toward greater transparency and human agency, externalizing a model's inferences in the natural language people already use to express what they want and letting them edit those inferences in real time. It also provides a structural lever for more targeted generalization, confining learning to a small, domain-relevant basis of dimensions. Together, they offer a proof of concept for an alignment paradigm in which models draw on their world knowledge of human preferences to make more ``reasonable'' inferences while being more transparent and steerable.

\bibliographystyle{plainnat}
\bibliography{references}


\appendix

\section{Discovered dimensions across domains}
\label{app:dimensions}

This appendix lists the full set of LLM-discovered preference dimensions deployed in each of the four domains covered in the paper. For \emph{moral dilemmas} (Table~\ref{tab:dims-moral-dilemmas}) and \emph{movies} (Table~\ref{tab:dims-movies}) we report the $K=10$ deployed basis used for the online studies, selected from a larger LLM-proposed pool by descending in-corpus projection variance (Section~\ref{sec:pipeline} and Appendix~\ref{app:dim-discovery}). For \emph{wines} (Table~\ref{tab:dims-wines}) and \emph{LLM responses} (Table~\ref{tab:dims-coalign}) we report the full $K=15$ basis used for offline method evaluation. The \textbf{Dimension} column is the display label produced by the pipeline's final labeling step, which automatically rewrites a handful of awkward auto-generated names for readability (e.g.\ ``Formality Avoidance''~$\rightarrow$~``Informality'') while preserving each dimension's polarity; these are the names shown to participants. All other fields---pole labels, descriptions, and ``typical person'' sketches---are reproduced verbatim from each run's \texttt{dimensions.json} file, with no manual editing or re-wording.

\begin{table}[H]
  \centering
  \footnotesize
  \caption{Full $K{=}10$ deployed basis for Moral dilemmas. Each row is one LLM-discovered dimension; per-pole label, description, and typical person are reproduced verbatim from \texttt{dimensions.json}.}
  \label{tab:dims-moral-dilemmas}
  \begin{tabularx}{\linewidth}{@{}>{\raggedright\arraybackslash}p{2.4cm} >{\raggedright\arraybackslash}X >{\raggedright\arraybackslash}X@{}}
    \toprule
    \textbf{Dimension} & \textbf{Low pole} & \textbf{High pole} \\
    \midrule
    \textbf{Moral Integrity} & ``Low Integrity'': Actions that may compromise personal ethics or values for convenience or gain. \emph{(Someone who prioritizes outcomes over ethical consistency.)} & ``High Integrity'': Actions that align closely with personal ethics, honesty, and moral consistency. \emph{(Someone who values ethical behavior and personal responsibility.)} \\
    \addlinespace[2pt]
    \textbf{Social Harmony} & ``Individualistic'': Actions that prioritize personal goals over group cohesion. \emph{(Someone who values personal autonomy over social norms.)} & ``Collectivist'': Actions that prioritize group cohesion, community standards, and social norms. \emph{(Someone who values social harmony and shared values.)} \\
    \addlinespace[2pt]
    \textbf{Informality} & ``Formal'': Actions that involve legal, institutional, or bureaucratic processes. \emph{(Someone who prefers structured and official procedures.)} & ``Informal'': Actions that avoid legal or formal processes in favor of personal or direct solutions. \emph{(Someone who prefers simplicity and direct action.)} \\
    \addlinespace[2pt]
    \textbf{Tradition Adherence} & ``Innovative'': Actions that challenge or deviate from established norms and traditions. \emph{(Someone who values creativity and change.)} & ``Traditional'': Actions that align with established norms, values, and customs. \emph{(Someone who values stability and continuity.)} \\
    \addlinespace[2pt]
    \textbf{Efficiency} & ``Inefficient'': Actions that are slow, complex, or resource-intensive. \emph{(Someone who values thoroughness over speed.)} & ``Efficient'': Actions that are quick, practical, and resource-conscious. \emph{(Someone who values speed and convenience.)} \\
    \addlinespace[2pt]
    \textbf{Privacy} & ``Public'': Actions that involve public exposure or social visibility. \emph{(Someone who is comfortable with public attention.)} & ``Private'': Actions that preserve personal boundaries and avoid public scrutiny. \emph{(Someone who values personal space and discretion.)} \\
    \addlinespace[2pt]
    \textbf{Financial Prudence} & ``Spontaneous Spending'': Actions that prioritize immediate financial gratification over long-term planning. \emph{(Someone who values immediate pleasure over financial security.)} & ``Financially Prudent'': Actions that prioritize long-term financial stability and responsibility. \emph{(Someone who values financial security and planning.)} \\
    \addlinespace[2pt]
    \textbf{Decisiveness} & ``Hesitant'': Actions that delay or avoid making a decision. \emph{(Someone who prefers to gather more information before acting.)} & ``Decisive'': Actions that involve prompt and confident decision-making. \emph{(Someone who values action and quick resolution.)} \\
    \addlinespace[2pt]
    \textbf{Creativity} & ``Conventional'': Actions that follow standard or expected approaches. \emph{(Someone who values structure and predictability.)} & ``Creative'': Actions that involve originality, self-expression, and innovation. \emph{(Someone who values imagination and artistic freedom.)} \\
    \addlinespace[2pt]
    \textbf{Structure Preference} & ``Flexible'': Actions that allow for adaptability and spontaneity. \emph{(Someone who values flexibility and change.)} & ``Structured'': Actions that follow predictable, organized, and planned approaches. \emph{(Someone who values order and routine.)} \\
    \bottomrule
  \end{tabularx}
\end{table}

\begin{table}[H]
  \centering
  \footnotesize
  \caption{Full $K{=}10$ deployed basis for Movies. Each row is one LLM-discovered dimension; per-pole label, description, and typical viewer are reproduced verbatim from \texttt{dimensions.json}.}
  \label{tab:dims-movies}
  \begin{tabularx}{\linewidth}{@{}>{\raggedright\arraybackslash}p{2.4cm} >{\raggedright\arraybackslash}X >{\raggedright\arraybackslash}X@{}}
    \toprule
    \textbf{Dimension} & \textbf{Low pole} & \textbf{High pole} \\
    \midrule
    \textbf{Emotional Depth} & ``Emotionally Neutral'': Minimal character development or emotional engagement. \emph{(Viewers prioritizing plot over character arcs.)} & ``Highly Emotional'': Rich character development and emotional resonance. \emph{(Viewers seeking transformative storytelling.)} \\
    \addlinespace[2pt]
    \textbf{Action Intensity} & ``Low Action'': Minimal physical conflict or kinetic sequences. \emph{(Viewers preferring dialogue-driven narratives.)} & ``High-Intensity Action'': Frequent, fast-paced physical conflict sequences. \emph{(Viewers seeking adrenaline-driven entertainment.)} \\
    \addlinespace[2pt]
    \textbf{Humor Intensity} & ``Not Humorous'': Minimal comedic elements or lighthearted moments. \emph{(Viewers avoiding slapstick or situational comedy.)} & ``Very Humorous'': Prominent comedic elements and lighthearted tone. \emph{(Viewers seeking entertainment through laughter.)} \\
    \addlinespace[2pt]
    \textbf{Historical Authenticity} & ``Modernized'': Contemporary reinterpretations or loose historical references. \emph{(Viewers prioritizing entertainment over accuracy.)} & ``Highly Authentic'': Detailed period accuracy in setting, language, and customs. \emph{(History enthusiasts and detail-oriented viewers.)} \\
    \addlinespace[2pt]
    \textbf{Suspense/Atmosphere} & ``Low Tension'': Minimal suspenseful moments or atmospheric build-up. \emph{(Viewers avoiding anxiety-inducing content.)} & ``Highly Suspenseful'': Pervasive tension, atmosphere, and anticipation. \emph{(Viewers seeking immersive, edge-of-seat experiences.)} \\
    \addlinespace[2pt]
    \textbf{Sci-Fi/Fantasy Worldbuilding} & ``Minimal Worldbuilding'': Basic or contemporary setting with few speculative elements. \emph{(Viewers preferring grounded narratives.)} & ``Rich Worldbuilding'': Detailed speculative worlds with unique rules and systems. \emph{(Viewers seeking immersive imaginative experiences.)} \\
    \addlinespace[2pt]
    \textbf{Survival/Stress Scenarios} & ``Low Stress'': Minimal life-threatening situations or survival challenges. \emph{(Viewers avoiding high-pressure narratives.)} & ``High-Stress Survival'': Centralized life-threatening challenges and survival stakes. \emph{(Viewers seeking intense survival narratives.)} \\
    \addlinespace[2pt]
    \textbf{Visual Spectacle} & ``Minimal Visuals'': Straightforward cinematography with few special effects. \emph{(Viewers prioritizing story over visuals.)} & ``High Visual Impact'': Immersive cinematography and prominent special effects. \emph{(Viewers appreciating visual artistry.)} \\
    \addlinespace[2pt]
    \textbf{Adventure Scope} & ``Local Scope'': Minimal exploration or journey-based elements. \emph{(Viewers preferring contained narratives.)} & ``Grand Adventure'': Epic journeys, exploration, and expansive settings. \emph{(Viewers seeking expansive adventure narratives.)} \\
    \addlinespace[2pt]
    \textbf{Family-Friendly Content} & ``Adult-Oriented'': Minimal consideration for younger audiences or family themes. \emph{(Viewers avoiding content for children.)} & ``Strong Family-Friendly'': Explicitly designed for family audiences with appropriate content. \emph{(Viewers seeking content for all ages.)} \\
    \bottomrule
  \end{tabularx}
\end{table}

\begin{table}[H]
  \centering
  \footnotesize
  \caption{Full $K{=}15$ basis for Wines (offline method evaluation). Each row is one LLM-discovered dimension; per-pole label, description, and typical drinker are reproduced verbatim from \texttt{dimensions.json}.}
  \label{tab:dims-wines}
  \begin{tabularx}{\linewidth}{@{}>{\raggedright\arraybackslash}p{2.4cm} >{\raggedright\arraybackslash}X >{\raggedright\arraybackslash}X@{}}
    \toprule
    \textbf{Dimension} & \textbf{Low pole} & \textbf{High pole} \\
    \midrule
    \textbf{Fruit Intensity} & ``Low fruit'': Minimal or no detectable fruit character; more neutral or earthy. \emph{(Someone who prefers savory or mineral-driven wines.)} & ``High fruit'': Prominent, ripe, and forward fruit flavors and aromas. \emph{(Someone who enjoys bold, fruit-forward reds or tropical/citrus whites.)} \\
    \addlinespace[2pt]
    \textbf{Body and Structure} & ``Light-bodied'': Delicate, low tannin, low alcohol, and low intensity on the palate. \emph{(Someone who prefers easy-drinking, sessionable wines.)} & ``Full-bodied'': Rich, weighty, and intense on the palate with high tannin or alcohol. \emph{(Someone who enjoys structured, bold reds or rich whites.)} \\
    \addlinespace[2pt]
    \textbf{Acidity} & ``Low acidity'': Smooth, flat, or flabby mouthfeel with little zing or freshness. \emph{(Someone who prefers softer, rounder wines.)} & ``High acidity'': Bright, crisp, and refreshing with a lively mouthfeel. \emph{(Someone who enjoys unoaked whites or high-acid reds.)} \\
    \addlinespace[2pt]
    \textbf{Oak Influence} & ``Unoaked'': Fresh, clean, and unadulterated by oak aging or barrel fermentation. \emph{(Someone who prefers crisp, natural, or modern styles.)} & ``Oaked'': Smoky, buttery, or vanilla notes from oak aging or barrel fermentation. \emph{(Someone who enjoys aged or complex reds or oaked whites.)} \\
    \addlinespace[2pt]
    \textbf{Sweetness} & ``Dry'': No perceptible residual sugar; clean and crisp finish. \emph{(Someone who prefers traditional, food-friendly wines.)} & ``Sweet'': Noticeable residual sugar with a sweet or off-dry finish. \emph{(Someone who enjoys dessert wines or sweet whites.)} \\
    \addlinespace[2pt]
    \textbf{Aromatic Complexity} & ``Simple nose'': Limited or straightforward aromas with little nuance. \emph{(Someone who prefers direct, easy-to-understand wines.)} & ``Complex nose'': Layered, evolving, and nuanced aromas with multiple dimensions. \emph{(Someone who enjoys aged or terroir-driven wines.)} \\
    \addlinespace[2pt]
    \textbf{Earthy/Savory Character} & ``Fruity focus'': Minimal earthy or savory notes; more fruit-driven and clean. \emph{(Someone who prefers bright, fruit-forward wines.)} & ``Earthy/savory'': Prominent notes of soil, herbs, mushrooms, or umami-like flavors. \emph{(Someone who enjoys rustic, food-friendly reds or aged whites.)} \\
    \addlinespace[2pt]
    \textbf{Aging Potential} & ``Drink now'': Meant to be consumed young; lacks structure or tannin for aging. \emph{(Someone who prefers approachable, early-drinking wines.)} & ``Aging potential'': High tannin, acidity, or alcohol that allows for long-term cellaring. \emph{(Someone who enjoys structured, collectible wines.)} \\
    \addlinespace[2pt]
    \textbf{Traditional Style} & ``Modern style'': Clean, fruit-forward, and made with modern techniques. \emph{(Someone who prefers accessible, contemporary wines.)} & ``Traditional style'': Old-world, terroir-driven, and made with minimal intervention. \emph{(Someone who values heritage and authenticity.)} \\
    \addlinespace[2pt]
    \textbf{Value for Money} & ``Premium price'': High cost relative to perceived quality or enjoyment. \emph{(Someone who prioritizes quality over cost.)} & ``Good value'': High quality relative to price; approachable and satisfying. \emph{(Someone who prioritizes affordability and enjoyment.)} \\
    \addlinespace[2pt]
    \textbf{Effervescence} & ``Still wine'': No bubbles or carbonation; smooth and still. \emph{(Someone who prefers traditional reds or whites.)} & ``Sparkling'': Perfected bubbles with a lively, effervescent mouthfeel. \emph{(Someone who enjoys celebratory or refreshing sparkling wines.)} \\
    \addlinespace[2pt]
    \textbf{Playfulness} & ``Serious style'': Formal, traditional, and conventional in presentation and branding. \emph{(Someone who values classic, elegant wines.)} & ``Playful style'': Unconventional, whimsical, or experimental in branding or presentation. \emph{(Someone who enjoys novelty or quirky wines.)} \\
    \addlinespace[2pt]
    \textbf{Tropical Aroma Intensity} & ``Non-tropical'': No detectable tropical fruit aromas; more citrus or stone fruit. \emph{(Someone who prefers traditional or European-style whites.)} & ``Tropical'': Prominent aromas of pineapple, mango, or passionfruit. \emph{(Someone who enjoys New World or warm-climate whites.)} \\
    \addlinespace[2pt]
    \textbf{Sessionability} & ``High alcohol'': Full-bodied, high ABV, and less suited for drinking in quantity. \emph{(Someone who enjoys bold, intense wines.)} & ``Low alcohol'': Light-bodied, low ABV, and easy to drink in quantity. \emph{(Someone who enjoys casual, everyday wines.)} \\
    \addlinespace[2pt]
    \textbf{Floral Aroma Intensity} & ``Non-floral'': No detectable floral notes; more fruit or earthy character. \emph{(Someone who prefers bold or savory wines.)} & ``Floral'': Prominent aromas of rose, lavender, or jasmine. \emph{(Someone who enjoys aromatic whites or ros\'es.)} \\
    \bottomrule
  \end{tabularx}
\end{table}

\begin{table}[H]
  \centering
  \footnotesize
  \caption{Full $K{=}15$ basis for LLM responses (offline method evaluation). Each row is one LLM-discovered dimension; per-pole label, description, and typical person are reproduced verbatim from \texttt{dimensions.json}.}
  \label{tab:dims-coalign}
  \begin{tabularx}{\linewidth}{@{}>{\raggedright\arraybackslash}p{2.4cm} >{\raggedright\arraybackslash}X >{\raggedright\arraybackslash}X@{}}
    \toprule
    \textbf{Dimension} & \textbf{Low pole} & \textbf{High pole} \\
    \midrule
    \textbf{Conciseness} & ``Wordy'': Long-winded, with excessive elaboration and tangents. \emph{(Someone who prefers depth over brevity.)} & ``Concise'': Direct, to the point, and free of unnecessary details. \emph{(Someone who values efficiency and time-saving.)} \\
    \addlinespace[2pt]
    \textbf{Structure} & ``Unstructured'': Chaotic, with no clear organization or logical flow. \emph{(Someone who prefers intuitive or narrative formats.)} & ``Structured'': Well-organized, with clear sections, headings, and logical progression. \emph{(Someone who values clarity and step-by-step guidance.)} \\
    \addlinespace[2pt]
    \textbf{Actionability} & ``Abstract'': Theoretical or general, with no clear steps to follow. \emph{(Someone who values reflection over action.)} & ``Actionable'': Provides clear, practical steps or recommendations. \emph{(Someone who values practical advice and implementation.)} \\
    \addlinespace[2pt]
    \textbf{Clarity} & ``Unclear'': Ambiguous, jargon-heavy, or difficult to understand. \emph{(Someone who prefers deep or poetic language.)} & ``Clear'': Straightforward, easy to understand, and free of ambiguity. \emph{(Someone who values direct and accessible communication.)} \\
    \addlinespace[2pt]
    \textbf{Emotional Resonance} & ``Detached'': Factual and impersonal, with little emotional engagement. \emph{(Someone who values logic and efficiency over emotion.)} & ``Emotionally Engaging'': Expressive, empathetic, and capable of evoking feelings. \emph{(Someone who values connection, inspiration, and emotional support.)} \\
    \addlinespace[2pt]
    \textbf{Depth} & ``Superficial'': Shallow, with little exploration of underlying ideas or implications. \emph{(Someone who values brevity and practicality.)} & ``In-Depth'': Rich in detail, nuance, and exploration of multiple perspectives. \emph{(Someone who values intellectual engagement and complexity.)} \\
    \addlinespace[2pt]
    \textbf{Descriptiveness} & ``Minimalist'': Sparse in imagery, tone, and sensory detail. \emph{(Someone who values clarity and efficiency.)} & ``Descriptive'': Rich in imagery, tone, and sensory detail, creating vivid mental images. \emph{(Someone who values immersive and expressive language.)} \\
    \addlinespace[2pt]
    \textbf{Sustainability} & ``Unaware'': No mention of environmental or ethical considerations. \emph{(Someone who prioritizes convenience or cost over sustainability.)} & ``Sustainable'': Emphasizes environmental responsibility, ethical practices, and long-term impact. \emph{(Someone who values eco-consciousness and social responsibility.)} \\
    \addlinespace[2pt]
    \textbf{Community Focus} & ``Individualistic'': Focuses on personal benefit or experience, with little regard for others. \emph{(Someone who values personal growth and self-improvement.)} & ``Community-Oriented'': Emphasizes shared experiences, social impact, and collective benefit. \emph{(Someone who values inclusivity, shared experiences, and social good.)} \\
    \addlinespace[2pt]
    \textbf{Historical Context} & ``Present-Focused'': No reference to historical background or cultural context. \emph{(Someone who values immediacy and practicality.)} & ``Historically Rich'': Includes detailed historical background, cultural context, and interpretive analysis. \emph{(Someone who values depth and cultural understanding.)} \\
    \addlinespace[2pt]
    \textbf{Practicality} & ``Theoretical'': Abstract or idealistic, with little regard for real-world application. \emph{(Someone who values intellectual exploration over implementation.)} & ``Practical'': Grounded in real-world application, with clear relevance to everyday life. \emph{(Someone who values actionable and accessible solutions.)} \\
    \addlinespace[2pt]
    \textbf{Efficiency} & ``Time-Consuming'': Requires significant time or effort to process or implement. \emph{(Someone who values depth and thoroughness over speed.)} & ``Efficient'': Quick to process, easy to implement, and time-saving. \emph{(Someone who values speed and minimal effort.)} \\
    \addlinespace[2pt]
    \textbf{Creativity} & ``Formulaic'': Predictable, conventional, and lacking originality. \emph{(Someone who values clarity and structure over novelty.)} & ``Creative'': Original, imaginative, and expressive, with unique perspectives or approaches. \emph{(Someone who values artistic expression and innovation.)} \\
    \addlinespace[2pt]
    \textbf{Authenticity} & ``Generic'': Standardized, impersonal, and lacking in personal or cultural nuance. \emph{(Someone who values efficiency and consistency over uniqueness.)} & ``Authentic'': Personal, culturally rich, and reflective of real-world experiences or traditions. \emph{(Someone who values cultural depth and real-world relevance.)} \\
    \addlinespace[2pt]
    \textbf{Formality} & ``Casual'': Informal, conversational, and relaxed in tone and structure. \emph{(Someone who values accessibility and relatability.)} & ``Formal'': Polished, professional, and structured in tone and presentation. \emph{(Someone who values professionalism and clarity.)} \\
    \bottomrule
  \end{tabularx}
\end{table}

\section{Feedback prior: construction and rescaling}
\label{app:rescaling}

This appendix specifies how the general feedback construction of Section~\ref{sec:inference} was instantiated in the two online studies (moral dilemmas and movies). In both, the discovered basis has $K=10$ dimensions and the top $M=3$ by the per-dimension--normalized projection $|u_{t,k}|/m_k$ are surfaced on each inference trial (Section~\ref{sec:behav-design}). Every step below is implemented identically in the offline Python fitter (\texttt{experiments/analyze.py}) and the in-browser JavaScript fitter (\texttt{web-interface/index.html}) (Appendix~\ref{app:implementation}).

\paragraph{Categories and verb phrases.} We instantiate the ordered, quantile-defined categories of Section~\ref{sec:inference} as $C=5$ levels per dimension, each carrying a domain-specific verb phrase $\eta^{(q)}$ that supplies its valence and intensity. For moral dilemmas the five phrases are \emph{downplay}, \emph{underweight}, \emph{are neutral about}, \emph{value}, and \emph{prioritize}; for movies they are \emph{prefer to skip}, \emph{aren't into}, \emph{are indifferent to}, \emph{like}, and \emph{love}. An inference renders the selected level's phrase with the dimension's name, e.g., ``You prioritize moral integrity'' or ``You aren't into action intensity.''

\paragraph{Bins and representative values.} Both the bin boundaries and the representative values $\theta_k^{(q)}$ are computed once per domain from the trial-pool distribution of per-dimension projections $u_{t,k} = v_k^\top \delta_t$, symmetrized by including each value together with its negation $-u_{t,k}$ so the categories are centered at $0$. For each dimension separately (the deployed \emph{per-dim} mode) we place the four bin boundaries at the $20,40,60,80$th percentiles of this distribution and the five representative values at the bin centers, the $10,30,50,70,90$th percentiles. By symmetry each bin then carries $\approx 20\%$ of the mass, the middle category's value is exactly $0$, and the table is antisymmetric ($\theta_k^{(1)} = -\theta_k^{(5)}$, $\theta_k^{(2)} = -\theta_k^{(4)}$). This per-dimension quintile-midpoint table is the \textsc{quintile\_midpoints} \emph{multiplier scheme}, our deployed default; alternative spacings (uniform across dimensions, or linear in the per-dim standard deviation) are swept in Appendix~\ref{app:calibration}.

\paragraph{From responses to the raw prior $\bar\theta_{\text{raw}}$.} On each inference trial and each of the $M=3$ visible dimensions, we quantize the chosen option's signed projection $(2z_t - 1)\,u_{t,k}$ into its quintile bin $q^\star$ and pre-select the corresponding category. The two conditions differ only in how the participant may revise this pre-selection:
\begin{itemize}
  \item In \textsc{inference\_affirm}, the participant \emph{affirms} the pre-selected category (retaining its representative value $\theta_k^{(q^\star)}$) or \emph{removes} the inference (contributing $0$).
  \item In \textsc{inference\_categories}, the participant selects any of the five levels (the pre-selection is shown highlighted as a suggestion), and the selected level's representative value is used.
\end{itemize}
Each visible (trial, dimension) pair thus contributes one value ($0$ if removed, otherwise the selected category's representative value) and we set $\bar\theta_{\text{raw}}[k]$ to the mean of these contributions over all trials on which dimension $k$ was visible; dimensions never surfaced to the participant remain $0$.

\paragraph{Rescaling to $\bar\theta$.} The raw prior $\bar\theta_{\text{raw}}$ lives on the scale of the per-trial projections $u_t$, while the choice-derived $\hat\theta$ lives on the larger scale of the BTL logit; left unrescaled, a given $\mu$ would impose almost no shrinkage. We therefore use a two-stage fit: (i) solve \eqref{eq:objective} with $\mu=0$ to obtain the choice-only fit $\hat\theta_{\text{data}}$; (ii) match the prior's norm to it,
\begin{equation}
  \bar\theta \;=\; \frac{\|\hat\theta_{\text{data}}\|}{\|\bar\theta_{\text{raw}}\|}\,\bar\theta_{\text{raw}};
  \label{eq:rescale}
\end{equation}
and (iii) refit \eqref{eq:objective} with $\mu>0$. This makes $\mu$ a scale-free trust ratio between feedback and choices rather than a raw magnitude, and renders the optimal $\mu$ approximately scale-invariant across domains. The values deployed at runtime were $\mu = 2.0,\ \lambda = 0.01$ (dilemmas) and $\mu = 1.0,\ \lambda = 0.005$ (movies); the body's H1 instead selects $(\text{scheme}, \mu, \lambda)$ by nested cross-validation (Section~\ref{sec:behavioral}), with the at-deployed-parameter analysis in Appendix~\ref{app:deployed-results}.

\section{Results at experimentally deployed parameters}
\label{app:deployed-results}

The body figure (Figure~\ref{fig:behav-main}) and table (Table~\ref{tab:behav-main}) report H1 under nested participant-level cross-validation: hyperparameters are picked on outer-train participants and the held-out lift is pooled across outer-test participants. This appendix reports the parallel analysis at the parameters that were actually deployed at the moment of data collection, namely the values the live Qualtrics interface used to render summaries and predictions during the experiment. These values are immutable: dilemmas $\mu = 2.0$, $\lambda = 0.01$; movies $\mu = 1.0$, $\lambda = 0.005$. They were chosen to produce conservative live previews (heavier shrinkage toward the inference prior) and were not optimized for held-out predictive accuracy.

\begin{figure}[H]
  \centering
  \includegraphics[width=0.98\linewidth]{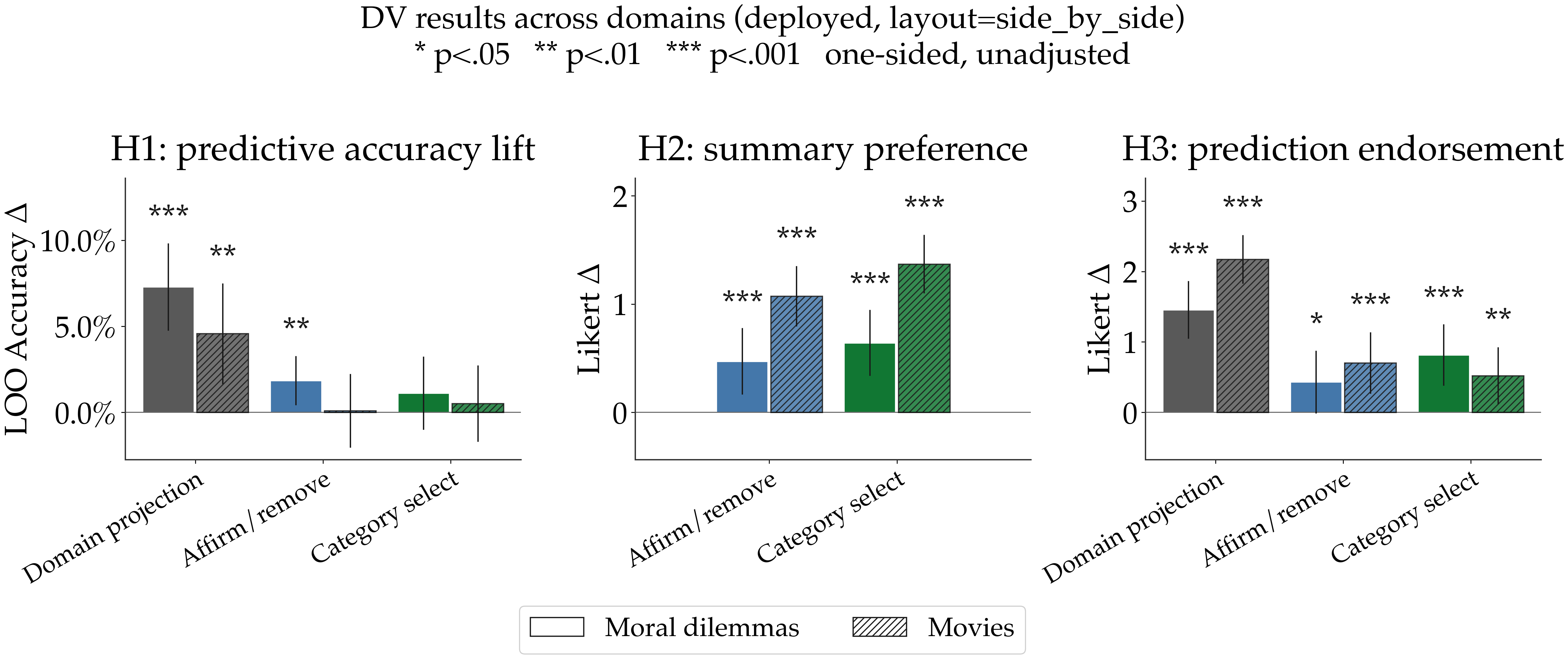}
  \caption{Behavioral results on moral dilemmas ($N=450$, solid bars) and movies ($N=449$, hatched bars) re-fit at the experimentally deployed parameters (dilemmas $\mu=2.0$, $\lambda=0.01$; movies $\mu=1.0$, $\lambda=0.005$). Compare with Figure~\ref{fig:behav-main} for the nested-CV version. The \textsc{choice\_only} contrast (which has no $\mu$ to tune, only a fixed $\lambda$) changes little on dilemmas and drops somewhat on movies relative to the nested-CV version; the inference conditions show smaller H1 lifts at the deployed conservative $\mu$, especially on movies where the deployed $\mu = 1.0$ over-shrinks toward the feedback prior. H2 and H3 are identical to the body figure (participant judgments of summaries and predictions rendered at exactly these deployed parameters).}
  \label{fig:behav-deployed}
\end{figure}

\begin{table}[H]
  \centering
  \small
  \begin{tabular*}{\linewidth}{@{\extracolsep{\fill}}lrrrr@{}}
    \toprule
    Condition & $N$ & H1 ($\Delta$acc pp, $p_{\text{holm}}$) & H2 (mean, $p_{\text{holm}}$) & H3 (mean $\Delta$, $p_{\text{holm}}$) \\
    \midrule
    \multicolumn{5}{@{}l}{\emph{Moral dilemmas} (deployed $\mu=2.0$, $\lambda=0.01$)} \\
    \textsc{choice\_only}            & 153 & $+7.3$, $<\!.0001$ & \multicolumn{1}{c}{---} & $+1.46$, $<\!.0001$ \\
    \textsc{inference\_affirm}       & 146 & $+1.8$, $.012$     & $+0.47$, $.001$          & $+0.43$, $.025$ \\
    \textsc{inference\_categories}   & 151 & $+1.1$, $.149$     & $+0.64$, $.0001$         & $+0.81$, $.0005$ \\
    \midrule
    \multicolumn{5}{@{}l}{\emph{Movies} (deployed $\mu=1.0$, $\lambda=0.005$)} \\
    \textsc{choice\_only}            & 152 & $+4.6$, $.004$     & \multicolumn{1}{c}{---} & $+2.17$, $<\!.0001$ \\
    \textsc{inference\_affirm}       & 151 & $+0.1$, $.647$     & $+1.07$, $<\!.0001$      & $+0.70$, $.002$ \\
    \textsc{inference\_categories}   & 146 & $+0.5$, $.647$     & $+1.37$, $<\!.0001$      & $+0.52$, $.007$ \\
    \bottomrule
  \end{tabular*}
  \vspace{.25em}
  \caption{Behavioral results at experimentally deployed parameters, parallel to Table~\ref{tab:behav-main} but with H1 re-fit at the deployed $(\mu, \lambda)$ rather than the nested-CV-tuned values. H2 and H3 are identical to the body table. H1 lifts are reported in percentage points. At the deployed (conservative) $\mu$, the inference-condition H1 lifts are smaller than at the nested-CV-tuned values; in the movies study they shrink to a non-significant $+0.1$ pp (\textsc{inference\_affirm}) and a non-significant $+0.5$ pp (\textsc{inference\_categories}). The deployed parameters served as defaults for live preview rendering during the experiment and are reported here as the immutable as-deployed record.}
  \label{tab:behav-deployed}
\end{table}

The qualitative reading is unchanged: the LLM-discovered basis still produces a large, statistically significant lift over a same-rank random projection (\textsc{choice\_only}), and the augmented summaries and predictions are still preferred over baselines (H2, H3). What changes are the H1 inference-condition lifts. At the deployed $\mu$ values (chosen for conservative live preview, not for predictive accuracy), the inference conditions show small or null H1 effects, especially on movies, where the deployed $\mu = 1.0$ heavily over-shrinks toward the feedback prior on a basis whose dimensional scales differ from dilemmas'. Treating the multiplier scheme, $\mu$, and $\lambda$ as hyperparameters and selecting them honestly on held-out participants (Section~\ref{sec:behavioral}) recovers the larger, significant effects the calibration sweep predicts.

\section{Calibration analysis}
\label{app:calibration}

\begin{figure}[H]
  \centering
  \includegraphics[width=0.95\linewidth]{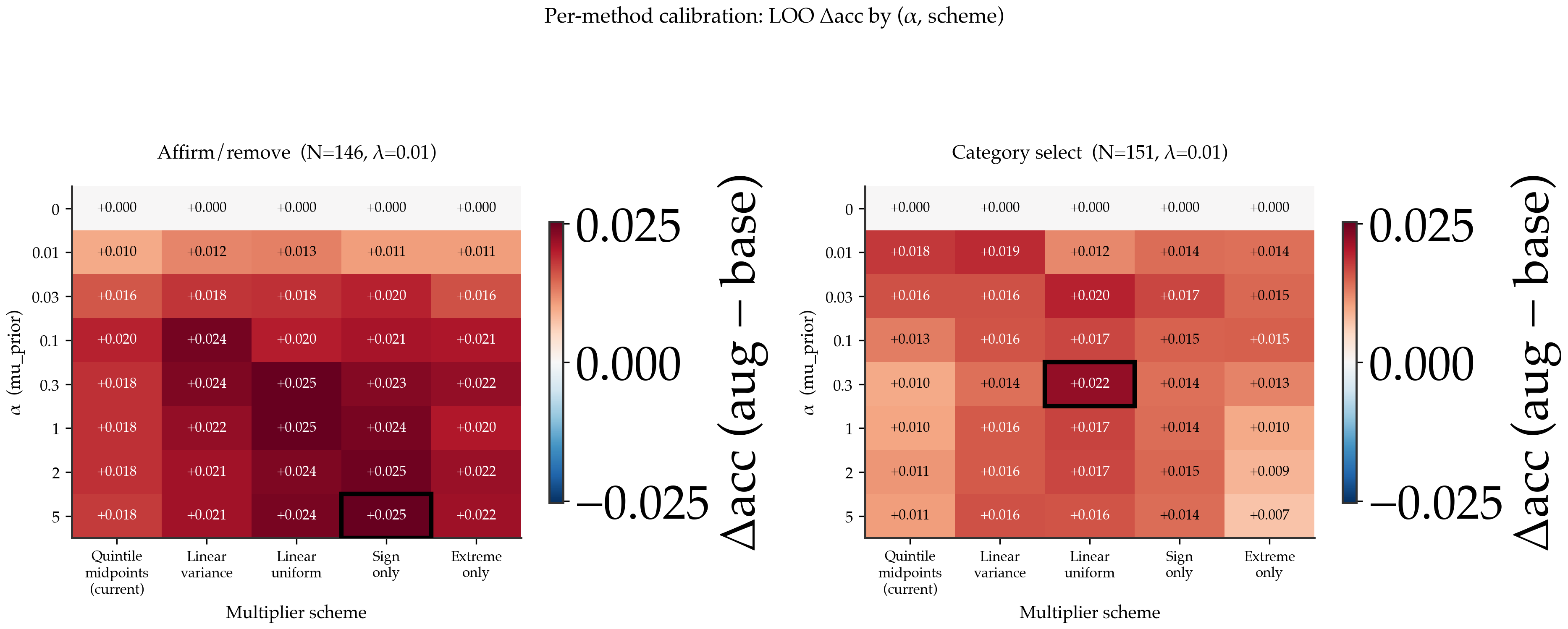}
  \caption{Calibration sweep on moral dilemmas ($N=450$). Each heatmap cell is the mean per-participant LOO accuracy lift (augmented $-$ projection-only baseline) at one $(\mu,\,\text{scheme})$ setting; rows are multiplier schemes, columns are $\mu$. The deployed setting is marked with a black outline; the per-condition optimum with a red outline. Left: \textsc{inference\_affirm}; right: \textsc{inference\_categories}.}
  \label{fig:calib-dilemmas}
\end{figure}

\begin{figure}[H]
  \centering
  \includegraphics[width=0.95\linewidth]{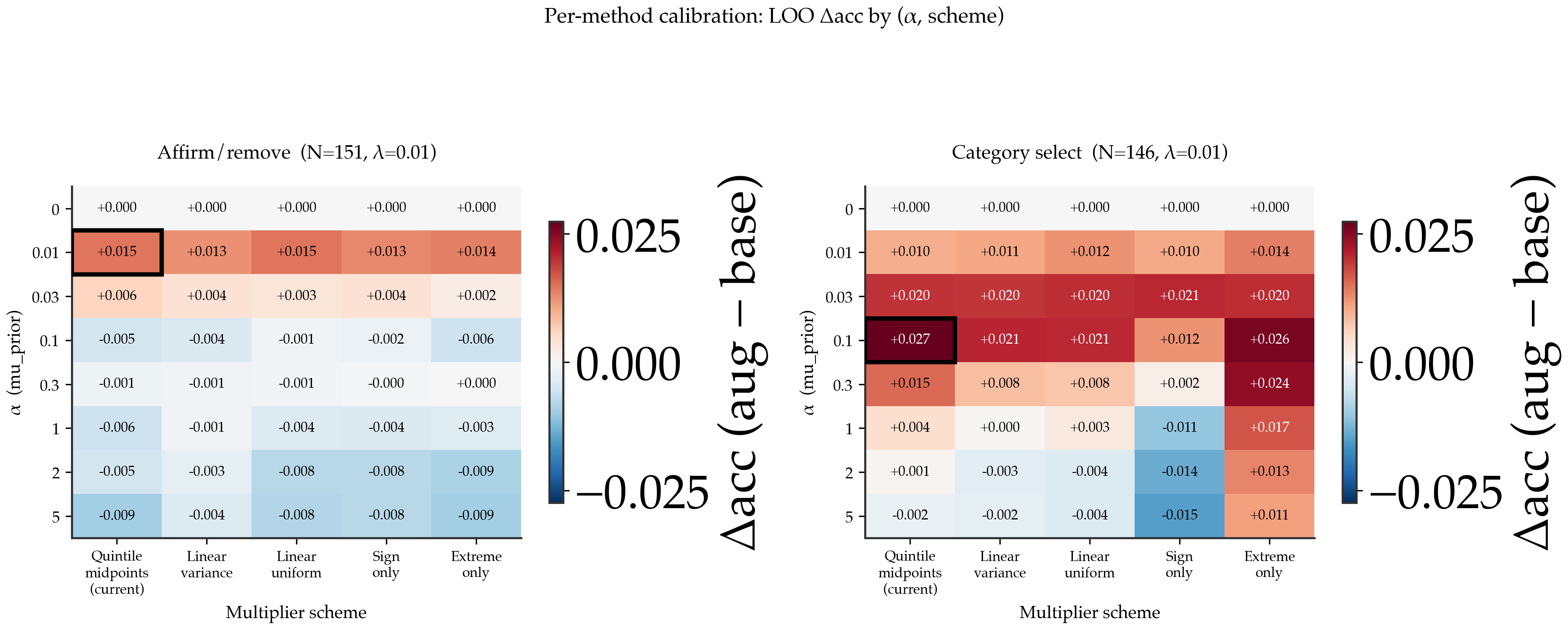}
  \caption{Calibration sweep on movies ($N=449$). Same axes as Figure~\ref{fig:calib-dilemmas}.}
  \label{fig:calib-movies}
\end{figure}

For each inference condition we swept the prior strength $\mu$ over $\{0, 0.01, 0.03, 0.1, 0.3, 1.0, 2.0, 5.0\}$ (with $\lambda=0.01$ fixed) and the multiplier scheme that maps category labels to per-dimension prior values over five variants: \textsc{quintile\_midpoints} (variance-weighted, quintile-derived spacing; the deployed default), \textsc{linear\_variance} (variance-weighted, linear spacing), \textsc{linear\_uniform} (uniform across dimensions, linear spacing), \textsc{sign\_uniform} (uniform, ternary $\{-1,0,+1\}$), and \textsc{extreme\_uniform} (uniform, only the most extreme categories contribute). For each cell we computed the per-participant LOO accuracy lift (augmented $-$ projection-only baseline), aggregated across participants in that condition. Figures~\ref{fig:calib-dilemmas} and~\ref{fig:calib-movies} show the resulting heatmaps for moral dilemmas and movies. Both panels reveal a clear interior optimum and substantial sensitivity to the choice of $\mu$ and scheme, which justifies treating $\mu$ as a deployment-time hyperparameter rather than a fixed constant.

\paragraph{Runtime UI vs.\ in-sample optimum.} Table~\ref{tab:calib-deployed-vs-opt} contrasts the parameters that ran in the live Qualtrics UI for real-time preview rendering during the experiment (\emph{Runtime UI}; immutable, also the basis of Appendix~\ref{app:deployed-results}) with the in-sample per-condition argmax of the calibration sweep (\emph{Per-condition in-sample optimum}). The runtime UI values were chosen to give conservative, well-shrunk previews, not to maximize held-out predictive accuracy; the in-sample optima exhibit selection bias (each cell of the sweep was scored on the whole sample, then the maximum was reported), so they cannot be taken at face value. The body's analysis (Section~\ref{sec:behavioral}, Figure~\ref{fig:behav-main}, Table~\ref{tab:behav-main}) instead uses nested participant-level cross-validation: hyperparameters are picked on outer-train participants and the held-out lift is pooled across outer-test participants, which sits between the two values reported below.

\begin{table}[H]
  \centering
  \footnotesize
  \setlength{\tabcolsep}{3pt}
  \caption{Two reference configurations of the inference fitter. \emph{Runtime UI} is what the live Qualtrics page used for real-time summary and prediction rendering during the experiment (locked at study launch; immutable). \emph{In-sample optimum} is the calibration-sweep argmax $(\text{scheme}, \mu, \lambda)$ per condition; the ``$\Delta$acc'' shown here is the in-sample LOO lift at that cell in percentage points, which has selection bias and is an upper bound on the held-out value. The honest held-out values appear in Table~\ref{tab:behav-main}.}
  \label{tab:calib-deployed-vs-opt}
  \begin{tabular*}{\linewidth}{@{\extracolsep{\fill}}l l l l c c c@{}}
    \toprule
    Domain & Configuration & Condition & scheme & $\mu$ & $\lambda$ & $\Delta$acc (in-sample, pp) \\
    \midrule
    \multirow{3}{*}{\textit{Moral dilemmas}} & Runtime UI               & both                 & quintile\_midpoints & 2.0 & 0.01 & --- \\
                                             & In-sample optimum        & inference\_affirm    & sign\_uniform       & 5.0 & 0.01 & $+2.5$ \\
                                             & In-sample optimum        & inference\_categories & linear\_uniform     & 0.3 & 0.01 & $+2.2$ \\
    \midrule
    \multirow{3}{*}{\textit{Movies}}         & Runtime UI               & both                 & quintile\_midpoints & 1.0 & 0.005 & --- \\
                                             & In-sample optimum        & inference\_affirm    & quintile\_midpoints & 0.01 & 0.01 & $+1.5$ \\
                                             & In-sample optimum        & inference\_categories & quintile\_midpoints & 0.10 & 0.01 & $+2.7$ \\
    \bottomrule
  \end{tabular*}
\end{table}

\paragraph{Deployment recommendation for new domains.} For a new domain without prior calibration data, we recommend $\lambda = 0.01$ and the \textsc{quintile\_midpoints} scheme, with $\mu$ selected by a quick within-domain pilot ($N \approx 50$) plus the same nested-CV protocol used in the body. Of the schemes we tested, \textsc{linear\_uniform} and \textsc{quintile\_midpoints} were near-optimal in at least one of our two online domains; \textsc{sign\_uniform} and \textsc{extreme\_uniform} are good fallbacks at small $N$ when the per-dim quintile-midpoint estimates are noisy. The optimal $\mu$ varies more across domains than $\lambda$ does in our data, suggesting that calibration of $\mu$ is the higher-leverage knob.

\section{Synthetic-participant simulations}
\label{app:simulation}

To complement the human-subjects experiments, we ran preference-vector simulations of the full experimental procedure on both deployed bases. The simulator instantiates the generative model implicit in our inference procedure: a synthetic participant has a ground-truth preference vector on the discovered subspace, makes BTL-noisy choices, optionally provides per-trial natural-language feedback, and we then fit the regularized-BTL estimator of Section~\ref{sec:inference} to the resulting (choice, feedback) data and read off held-out accuracy and log-likelihood. Every simulation run uses the deployed hyperparameters of its domain ($\mu{=}2.0,\,\lambda{=}0.01$ for moral dilemmas; $\mu{=}1.0,\,\lambda{=}0.005$ for movies), $K{=}10$ dimensions (top-$10$ selected from each domain's full LLM-discovered basis by descending in-corpus projection variance, matching the deployed selection in Section~\ref{sec:setup}), top-$3$ visible dimensions per inference trial, and $50$ synthetic participants per condition. The simulation code is in \texttt{simulation/run\_simulation.py}.

\paragraph{Synthetic participants.} Each participant is parameterized by a sparse ground-truth weight vector $w^\star \in \mathbb{R}^K$. We sample 2--3 ``active'' dimensions per participant (target sparsity $\approx 0.3$), assign each active dimension a weight uniformly drawn from $\pm[1.5, 3.0]$, and set the remaining components to zero. This emulates participants who care strongly about a few specific qualities and are indifferent to the rest, which is consistent with what we observe in the human data, where most participants' fitted $\theta$ has substantial mass on a small handful of dimensions.

\paragraph{Trials and choices.} For each synthetic participant we sample $T=20$ trial pairs uniformly from the corpus and project each option onto the discovered basis to obtain $\phi_K(a) \in \mathbb{R}^K$. The participant's choice on trial $t$ is sampled from the BTL likelihood
\begin{equation*}
  \Pr(a \succ a' \mid w^\star) \;=\; \sigma\bigl(\beta \cdot {w^\star}^{\!\top}\, (\phi_K(a) - \phi_K(a'))\bigr),
\end{equation*}
with choice-temperature $\beta=2$. Each participant's true LOO accuracy ceiling is therefore controlled, not perfect at $T=20$, but well above chance.

\paragraph{Feedback simulation in inference conditions.} On each inference trial we identify the top-$3$ dimensions by $|V^\top \delta_t|$, a simplified analogue of the visible-dimension rule used in the human experiments (which ranked by the per-dimension--normalized projection $|u_{t,k}|/m_k$ of Section~\ref{sec:inference}; the simulator omits the per-dimension normalization). For each visible dimension $k$, the simulator computes the participant's ``true'' category (the per-dim quintile bin into which $w^\star_k$ falls under the symmetric multiplier set $\{-1.5,-1,0,+1,+1.5\}$) and the model's pre-selected category (the bin into which the trial's per-dim projection falls). \textsc{inference\_categories} participants then return their true category, perturbed by a per-dimension slip probability of $0.1$ that flips them to an adjacent bin. \textsc{inference\_affirm} participants return \textsc{affirm} or \textsc{remove}: they affirm when the model's pre-selected category has the correct sign and remove when the dimension is irrelevant to them or the pre-selection has the wrong sign, again with a 0.1 per-dim slip rate. The aggregated per-trial categorical responses are passed through the same midpoint-based prior construction (Section~\ref{sec:inference}) used on the real data.

\paragraph{Fitter, evaluation, and metrics.} We fit the same regularized BTL objective from Equation~\eqref{eq:objective} by Newton's method, with the deployed $(\mu, \lambda)$ for each domain. At every prefix length $t \in \{1, \ldots, T\}$, we (i) refit on the first $t$ trials and predict each held-out trial using leave-one-out cross-validation, (ii) record LOO accuracy and held-out log-likelihood for the random projection, the projection-only model, and the projection $+$ feedback prior model. Final-trial metrics roll up into the LOO summary in Table~\ref{tab:sim-loo}; checkpoint-by-checkpoint metrics drive the learning-curve figures below.

\begin{figure}[H]
  \centering
  \includegraphics[width=0.98\linewidth]{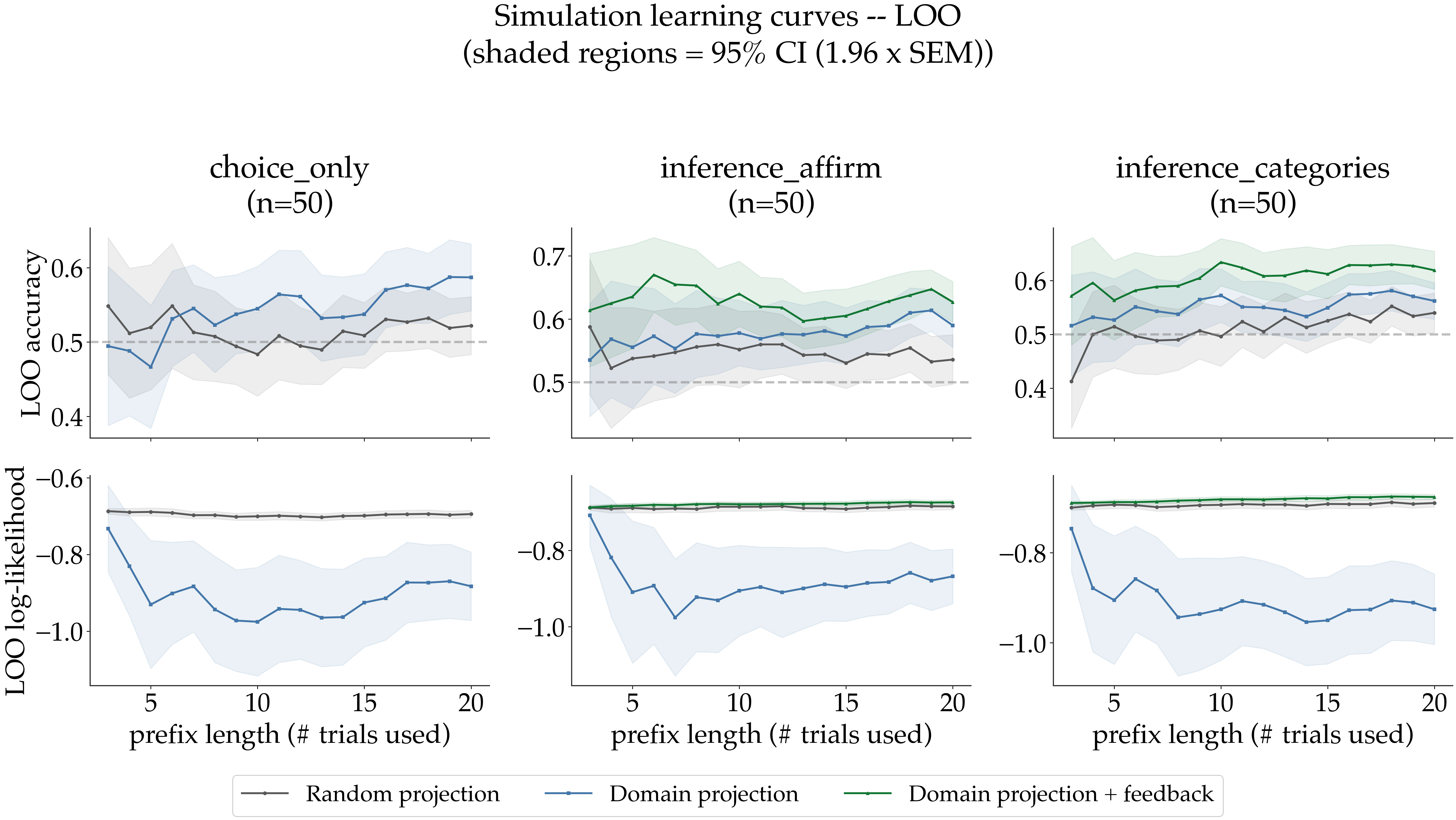}
  \caption{Simulation learning curves on moral dilemmas at deployed parameters ($\mu=2.0$, $\lambda=0.01$). Top row: held-out LOO accuracy as a function of prefix length, by condition; bottom row: held-out log-likelihood. Mean $\pm$ 95\% CI across 50 synthetic participants. Same gray/blue/green model encoding as the main-text figures.}
  \label{fig:sim-dilemmas-curves}
\end{figure}

\begin{figure}[H]
  \centering
  \includegraphics[width=0.98\linewidth]{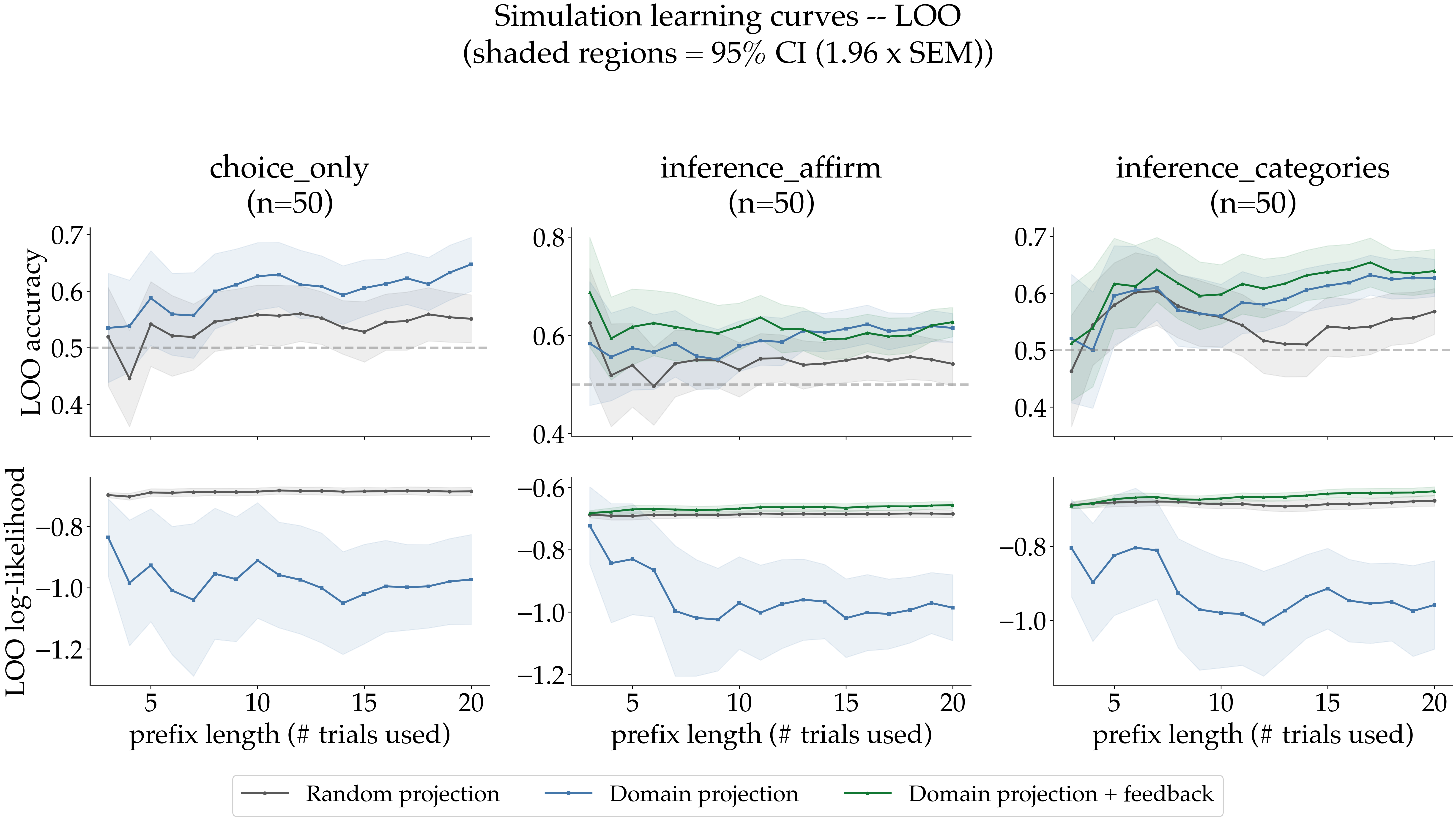}
  \caption{Simulation learning curves on movies at deployed parameters ($\mu=1.0$, $\lambda=0.005$). Same axes and conventions as Figure~\ref{fig:sim-dilemmas-curves}.}
  \label{fig:sim-movies-curves}
\end{figure}

\begin{table}[H]
  \centering
  \footnotesize
  \setlength{\tabcolsep}{3pt}
  \caption{Simulation LOO accuracy at $T=20$ across conditions, both domains, deployed hyperparameters. Accuracies are in percent; lift columns are in percentage points. ``Random'' is a same-rank random orthonormal projection; ``proj.\ only'' is the discovered basis with $\mu=0$; ``proj.\ +\,$\mu$'' adds the feedback prior. The lift columns ($\Delta$) parallel the H1 statistic in Table~\ref{tab:behav-main}.}
  \label{tab:sim-loo}
  \begin{tabular*}{\linewidth}{@{\extracolsep{\fill}}llrrrrr@{}}
    \toprule
    Domain & Condition & Random (\%) & proj.\ only (\%) & proj.+$\mu$ (\%) & $\Delta$ proj. vs.\ rand (pp) & $\Delta\,\mu$ vs.\ proj. (pp) \\
    \midrule
    Moral dilemmas & Domain projection     & 52.2 & 58.7 & 58.7 & $+6.5$ & $+0.0$ \\
                   & Affirm/remove         & 53.6 & 59.0 & 62.7 & $+5.4$ & $+3.7$ \\
                   & Category select       & 54.0 & 56.2 & 61.9 & $+2.2$ & $+5.7$ \\
    \midrule
    Movies         & Domain projection     & 55.1 & 64.7 & 64.7 & $+9.6$ & $+0.0$ \\
                   & Affirm/remove         & 54.2 & 61.5 & 62.7 & $+7.3$ & $+1.2$ \\
                   & Category select       & 56.8 & 62.7 & 63.9 & $+5.9$ & $+1.2$ \\
    \bottomrule
  \end{tabular*}
\end{table}

The simulations recover the main qualitative pattern in the human data: on both domains, the discovered basis improves LOO accuracy over a same-rank random projection by $+6.5$ percentage points on dilemmas and $+9.6$ on movies (domain-projection rows), the same direction as the empirical manipulation check ($+8.1$ pp on dilemmas, $+6.8$ pp on movies in Table~\ref{tab:behav-main}). The feedback prior gives a further lift in inference conditions, again consistent with the human data. Effects in the simulator are larger than in the human data, particularly for category-select on dilemmas (sim $+5.7$ pp vs.\ human held-out $+2.3$ pp in Table~\ref{tab:behav-main}). This may be because the synthetic participant gives noisier-than-uniform but otherwise calibrated feedback, while real participants exhibit additional structure (e.g., systematic over-/under-reporting of certain dimensions) that the multiplier scheme does not absorb perfectly, or because the simulator is run at the deployed runtime parameters of each domain, which the nested-CV inner search in Section~\ref{sec:behavioral} consistently moves \emph{away} from (toward smaller $\mu$ and smaller $\lambda$ on dilemmas). Calibration sensitivity to the multiplier scheme and prior strength is examined in Appendix~\ref{app:calibration}.

\section{Implementation details}
\label{app:implementation}

The objective in Equation~\eqref{eq:objective} is minimized by Newton's method with convergence tolerance $10^{-7}$ on the maximum coordinate update and a 15-iteration cap; convergence typically takes 5--10 iterations. The fit is fast enough to run in a participant's browser in real time, which we use to render augmented model summaries and held-out predictions during the post-task evaluation.

The Newton fit is implemented twice: in Python for offline analysis (\texttt{experiments/analyze.py}) and in JavaScript for the in-browser evaluation (\texttt{web-interface/index.html}). The two implementations share an identical loss function (Equation~\eqref{eq:objective}), Newton update rule, and convergence tolerance ($10^{-7}$ on the maximum coordinate update).

The two implementations also share the per-trial feedback-prior construction (the quintile-midpoint mapping that produces $\bar\theta_{\text{raw}}$, described in Appendix~\ref{app:rescaling}): the midpoint table, the action-to-midpoint mapping, and the averaging logic are written once in Python and once in JavaScript.

\paragraph{Random-projection baseline.} The baseline subspace $V_{\text{rand}}$ used in the \textsc{choice\_only} contrast is generated once, in \texttt{experiments/select\_top\_dims.py}, by taking the QR decomposition of a $d \times K$ matrix of i.i.d.\ standard-Gaussian entries and using the resulting orthonormal columns; this yields a rank-$K$ orthonormal basis matched to the discovered $V$. Generation uses a fixed seed (offset $+777$ from the pipeline seed), and the resulting projections $V_{\text{rand}}^\top \delta_t$ are precomputed and stored alongside the discovered projections in \texttt{trial\_projections.json}, so the \emph{same} $V_{\text{rand}}$ is shared across all participants rather than resampled per participant. This holds the $d \to K$ dimensionality reduction fixed between baseline and method, isolating the quality of the discovered directions from the effect of dimensionality reduction alone.

\paragraph{Reproducibility.} The full code release accompanying this paper (available at \url{https://github.com/Zachary-Wojtowicz/weights-to-words}) reproduces every result in three levels:
\begin{enumerate}
  \item \emph{Analysis only}: \texttt{python experiments/pipeline.py} regenerates every table and figure in Section~\ref{sec:behavioral} and the appendices from the cached participant data in \texttt{experiments/\{dilemmas,movies\}/data.csv} and the cached per-domain pipeline outputs.
  \item \emph{Analysis + simulations}: \texttt{python simulation/run\_simulation.py} on each domain regenerates Appendix~\ref{app:simulation}.
  \item \emph{Full pipeline from scratch}: per-domain bash scripts \texttt{run\_dailydilemmas.sh}, \texttt{run\_movies\_100.sh}, \texttt{run\_wines\_100.sh}, \texttt{run\_coalign\_50.sh} orchestrate the eight implementation steps (prepare $\to$ embed $\to$ select $\to$ discover dimensions $\to$ score options $\to$ fit BTL $\to$ find directions $\to$ evaluate) that realize the five-stage pipeline of Section~\ref{sec:pipeline} (the first three steps prepare the corpus for stage~1, and \emph{evaluate} produces the Section~\ref{sec:decomp} diagnostics). Each script is idempotent and skips steps whose outputs already exist. A vLLM endpoint serving Qwen3-Embedding-8B and an instruction-tuned LLM endpoint are required. End-to-end runtime is approximately 4 hours per domain on 4$\times$A100 GPUs (80 GB), dominated by the LLM-as-judge pairwise scoring stage; the analysis-only and analysis$+$simulation paths complete in minutes on a CPU.
\end{enumerate}

\paragraph{Data and code release.} The repository contains: (i) the participant data CSVs from both online studies (de-identified; only the Prolific-issued response IDs are retained, with no IPs, demographics that could re-identify, or free-text comments included in the public release); (ii) the per-domain pipeline outputs (\texttt{dimensions.json}, \texttt{bt\_scores.csv}, \texttt{directions.npz}, \texttt{trial\_projections.json}); (iii) the prompt templates listed in this appendix; (iv) the Qualtrics integration script \texttt{web-interface/qualtrics\_qid2\_questionjs.txt}; and (v) the consent form. All released under a permissive open-source license. The pre-registrations remain at the AsPredicted URLs cited in Section~\ref{sec:behavioral}.

\section{Dimension-discovery procedure}
\label{app:dim-discovery}

\paragraph{Pipeline LLMs and embedders.} The choice feature map $\phi$ is Qwen3-Embedding-8B (last-token pooling, $d=4096$);\footnote{\url{https://huggingface.co/Qwen/Qwen3-Embedding-8B}} the reason-elicitation, deduplication, dimension-condensation, and option-scoring stages all use a single instruction-tuned LLM (Qwen3-32B in all runs reported here). The coverage-validation stage and the embedding-clustering fallback embed reasons and pole descriptions with OpenAI's \texttt{text-embedding-3-small}.\footnote{\url{https://platform.openai.com/docs/guides/embeddings}}

\paragraph{Pipeline defaults.} Across all domains we use: $S=100$ option pairs sampled in three cosine-distance strata; $k=5$ reasons elicited per side at temperature $0.7$; ${\sim}50$ themes targeted in the deduplication stage; the top ${\approx}8K$ (eight per dimension) themes passed to the condenser; coverage-validation similarity threshold $0.4$; ${\sim}30$ pairwise judgments per option in the scoring stage, double-judged with order swap and filtered for swap consistency; ridge regularizer $\alpha_k$ in Equation~\eqref{eq:ridge} selected by leave-one-out CV from $\{10^{-2},10^{-1},1,10,10^2,10^3\}$. The resulting direction stack is L2-normalized row-wise (not orthogonalized); subspace coverage metrics in Section~\ref{sec:decomp} are evaluated on a QR-orthonormalization of $V$ that spans the same subspace. We use $K=10$ for moral dilemmas and movies and $K=15$ for wines and LLM responses, reflecting the elbow of the basis-coverage scree plot per domain.

\paragraph{Prompt templates.} The pipeline uses seven prompt templates across the four LLM-driven stages (reason elicitation has three switchable variants and option scoring has two), plus one additional template for the alternative direct-generation pipeline (\texttt{method\_llm\_gen/}). All templates are reproduced verbatim below; they are also released alongside the paper under \texttt{method\_llm\_examples/prompts/} (reason elicitation, deduplication, and dimension condensation) and \texttt{method\_llm\_gen/prompts/} (option scoring and the direct-generation variant). Variable substitutions (e.g., \texttt{\{domain\_items\}}, \texttt{\{reason\_list\}}, \texttt{\{N\}}) are filled in from the per-domain configuration at run time. The deployed pipeline uses \texttt{reason\_elicit.txt} for reason elicitation; the two variants (\texttt{values}, \texttt{persona}) target alternative framings of the same task and are included for completeness.

\subsubsection*{Reason elicitation.}
The default prompt asks the LLM to produce $k$ short, transferable preferences for each side of a candidate pair, without naming the specific options. This is the prompt used in every deployed run reported in the paper.

\noindent\texttt{method\_llm\_examples/prompts/reason\_elicit.txt}: default reason-elicitation prompt.
\begin{lstlisting}
Here are two {domain_items}. Imagine a large, diverse group of people each
choosing which one they prefer. Different people will choose differently --
and for different reasons.

{DOMAIN_ITEM} A:
{option_a_text}

{DOMAIN_ITEM} B:
{option_b_text}

List {reasons_per_side} distinct reasons someone might choose {DOMAIN_ITEM} A over B, and
{reasons_per_side} distinct reasons someone might choose B over A. Each reason should
identify a single preference or taste that makes one option more appealing
to a particular type of person.

IMPORTANT: State each reason as a SHORT, GENERAL preference -- one that would
apply across many different {domain_items}, not just this pair. The pair above
is a stimulus to help you think of preferences, but the reason itself should
be a transferable taste or priority. Do NOT mention the specific {domain_items}
by name in the reason.

State each reason as a SINGLE CONCEPT -- what the person is drawn to -- not
as a contrast between two things. The contrast is already captured by which
side (A or B) the reason appears on.

Good examples:
- "Preference for satirical, irreverent humor"
- "Drawn to slow-building atmospheric tension"
- "Values historical authenticity and period detail"

Bad examples:
- "Preference for humor over horror" (don't specify what it's over)
- "People who enjoy parodies of Sparta..." (too specific to this pair)
- "It has better acting" (quality judgment, not a preference)

Keep each reason to ONE sentence (under 15 words if possible). Avoid generic
quality judgments -- focus on what *type* of content or experience the person
is drawn to.

Format (plain text, not JSON):

REASONS TO CHOOSE A:
1. [reason]
2. [reason]
3. [reason]
4. [reason]
5. [reason]

REASONS TO CHOOSE B:
1. [reason]
2. [reason]
3. [reason]
4. [reason]
5. [reason]
\end{lstlisting}

The same task can be framed to surface preferences grounded in personal values or in rich personas. We provide both as switchable variants:

\noindent\texttt{method\_llm\_examples/prompts/reason\_elicit\_values.txt}: values-grounded variant.
\begin{lstlisting}
Here are two {domain_items}. Imagine a large, diverse group of people each
choosing which one they prefer. Different people will choose differently --
and for different reasons rooted in their personal values and life experiences.

{DOMAIN_ITEM} A:
{option_a_text}

{DOMAIN_ITEM} B:
{option_b_text}

List {reasons_per_side} distinct reasons someone might choose {DOMAIN_ITEM} A over B, and
{reasons_per_side} distinct reasons someone might choose B over A.

For each reason, describe the KIND OF PERSON who would have this preference
and WHY their values or life experience lead them to it. Each reason should
be 1-2 sentences starting with a phrase like "A person who..." or "Someone
with..." that grounds the preference in a recognizable value system or
background.

IMPORTANT: State each reason as a GENERAL preference -- one that would apply
across many different {domain_items}, not just this pair. The pair above is
a stimulus to help you think of preferences, but the reason itself should
be a transferable value or priority. Do NOT mention the specific {domain_items}
by name in the reason.

Good examples:
- "A person who prioritizes measurable and evidence-based outcomes may prefer an approach that directly addresses quantifiable metrics."
- "Someone with a history of chronic issues may be drawn to solutions that offer rapid, tangible improvements in their condition."

Bad examples:
- "This option is better because it has more detail" (quality judgment, not a value)
- "People who like Option A's specific approach to X" (too tied to this pair)

Format (plain text, not JSON):

REASONS TO CHOOSE A:
1. [reason]
2. [reason]
3. [reason]
4. [reason]
5. [reason]

REASONS TO CHOOSE B:
1. [reason]
2. [reason]
3. [reason]
4. [reason]
5. [reason]
\end{lstlisting}

\noindent\texttt{method\_llm\_examples/prompts/reason\_elicit\_persona.txt}: rich-persona variant.
\begin{lstlisting}
Here are two {domain_items}. Imagine a large, diverse group of people each
choosing which one they prefer. Different people will choose differently --
and for different reasons.

{DOMAIN_ITEM} A:
{option_a_text}

{DOMAIN_ITEM} B:
{option_b_text}

List {reasons_per_side} distinct reasons someone might choose {DOMAIN_ITEM} A over B, and
{reasons_per_side} distinct reasons someone might choose B over A.

For each reason, build a RICH PERSONA SKETCH of the person who holds this
preference. Describe their mindset, background, motivations, and what this
choice reveals about their deeper priorities. Each reason should be 2-3
sentences that paint a vivid picture of who this person is and why they
would make this choice.

IMPORTANT: State each reason as a GENERAL preference -- one that would apply
across many different {domain_items}, not just this pair. The pair above is
a stimulus to help you think of preferences, but the reason itself should
be a transferable trait or priority. Do NOT mention the specific {domain_items}
by name in the reason.

Good examples:
- "The person has a results-oriented mindset and is willing to tolerate short-term discomfort for measurable, data-backed improvements. This reflects a pragmatic worldview shaped by professional or academic environments."
- "They are drawn to holistic approaches, reflecting a belief in the body's innate ability to heal. This suggests an intuitive or spiritual orientation to wellbeing."

Bad examples:
- "This option is better" (quality judgment, not a persona)
- "People who like the first option" (not a rich persona sketch)

Format (plain text, not JSON):

REASONS TO CHOOSE A:
1. [reason]
2. [reason]
3. [reason]
4. [reason]
5. [reason]

REASONS TO CHOOSE B:
1. [reason]
2. [reason]
3. [reason]
4. [reason]
5. [reason]
\end{lstlisting}

\subsubsection*{Reason deduplication.}
Consolidates the elicited free-text reasons into a smaller set of recurring themes. We chunk the reason pool to fit the LLM's context window and merge results across batches.

\noindent\texttt{method\_llm\_examples/prompts/reason\_dedup.txt}: deduplication prompt (LLM-based clustering).
\begin{lstlisting}
Below are {N} reasons that people gave for choosing one {domain_item} over
another. Many reasons express the same underlying theme in different words.

Your task: identify the ~{target_themes} most frequently recurring distinct
themes across these reasons. For each theme, provide:
- A brief label (5-10 words) that captures the core preference
- A count of how many of the original reasons express this theme
- The IDs of the reasons that express this theme

Be aggressive about merging -- if two reasons reflect the same underlying
preference (even if applied to different specific options), they are the
same theme. But do NOT merge themes that are genuinely distinct preferences.

REASONS:
{reason_list}

Respond with valid JSON only -- no prose before or after, no markdown fences.

{{
  "themes": [
    {{
      "theme_id": 1,
      "label": "brief theme label",
      "count": 42,
      "reason_ids": [0, 3, 7, 12]
    }}
  ]
}}
\end{lstlisting}

\subsubsection*{Dimension condensation.}
Takes the top themes and produces $K$ unipolar named dimensions, each with low/high pole descriptions, articulability/variance hints, and the themes it subsumes (used downstream for coverage validation).

\noindent\texttt{method\_llm\_examples/prompts/dimension\_condense.txt}: condensation prompt; emits the deployed \texttt{dimensions.json}.
\begin{lstlisting}
You are analyzing reasons people give for choosing between {domain_items}.

Below is a deduplicated list of preference themes people expressed when
choosing between {domain_items}. The count indicates how many times each
theme appeared across {num_pairs} different choice pairs -- higher counts
indicate more pervasive preferences.

Your task is to identify the {K} most important UNIPOLAR preference features
that explain these themes. Each feature represents a single quality that a
{domain_item} can have MORE or LESS of -- not a contrast between two
different qualities.

Each dimension should:
- Be a UNIPOLAR FEATURE -- it describes a single quality that varies from
  "absent/minimal" to "strongly present" (e.g., "Humor" ranges from
  "no humor" to "very humorous")
- NOT be a forced contrast between two different concepts (e.g., do NOT
  create "Humor vs. Horror" -- instead create separate "Humor" and "Horror"
  features if both are important)
- SUBSUME as many of the listed themes as possible
- Be DISTINCT from the other dimensions -- two dimensions should not rank
  options in nearly the same order
- Be PREFERENCE-RELEVANT -- something that different people value differently,
  not a neutral descriptor

For each dimension, cite which theme numbers it subsumes (this verifies
coverage and coherence).

THEMES (sorted by frequency):
{reason_list}

Respond with valid JSON only -- no prose before or after, no markdown fences.

{schema}
\end{lstlisting}

\subsubsection*{Option scoring (LLM-as-judge).}
Two scoring variants exist. The deployed path is \emph{pairwise}: for each dimension we sample ${\sim}30$ pairwise judgments per option (each pair judged twice with order swapped, only swap-consistent judgments retained), then fit a Bradley--Terry model on the resulting wins/losses to recover scalar option scores $s_k(a_n) \in \mathbb{R}$. Pairwise judging is more robust to LLM scoring drift across options because every comparison is local; the BT fit aggregates across redundant judgments and handles transitivity automatically. The \emph{direct-scalar} variant asks the LLM to emit an integer score in $\{0,1,2,3,4,5\}$ per (dimension, option) directly. It is faster ($N$ calls per dimension instead of ${\sim}30 N$) but more sensitive to score-scale drift across the corpus; we include it here for completeness and use it as a smoke test during pipeline development. Both variants consume the dimension definitions (name, pole labels, scoring guidance, and example anchors) emitted by the dimension-condensation stage and feed the same downstream ridge-regression direction-fitting stage.

\noindent\texttt{method\_llm\_gen/prompts/llm\_binary\_judge.txt}: pairwise judging prompt.
\begin{lstlisting}
BINARY_CHOICE_PROMPT_TEMPLATE = """\
You are comparing two options on a single preference dimension.
Your task is to identify which option exhibits MORE of this quality,
or whether the difference is negligible.

--- DIMENSION ---
Name: {name}
Low end -- {low_label}: {low_description}
High end -- {high_label}: {high_description}
Scoring guidance: {scoring_guidance}

--- CALIBRATION ANCHORS ---
Example scoring low on this dimension: {low_example}
Example scoring high on this dimension: {high_example}

--- OPTIONS ---
Option A:
{option_a}

Option B:
{option_b}

--- TASK ---
Think step by step:
1. How much of this quality does Option A exhibit?
2. How much of this quality does Option B exhibit?
3. Which option exhibits more of it, and by how much?

Then record your final judgment.

--- RESPONSE FORMAT ---
Respond with valid JSON only, no prose before or after:
{{
  "reasoning": "<2-4 sentences comparing the two options on this dimension>",
  "choice": "A" | "B" | "negligible",
  "confidence": "clear" | "marginal",
  "high_pole_description_a": "<one phrase: how much Option A exhibits this quality>",
  "high_pole_description_b": "<one phrase: how much Option B exhibits this quality>"
}}

Use "negligible" only when the two options are genuinely indistinguishable on this
dimension -- not as a hedge. If you use "negligible", set confidence to "marginal".
"""
\end{lstlisting}

\noindent\texttt{method\_llm\_gen/prompts/llm\_score\_judge.txt}: direct-scalar scoring prompt. The LLM emits an integer 0--5 per option, with anchors taken from the same dimension definition.
\begin{lstlisting}
JUDGE_SCORING_PROMPT_TEMPLATE = """\
You are scoring one option on one preference dimension.

--- DIMENSION ---
Name: {name}
Score 0 -- {low_label}: {low_description}
Score 5 -- {high_label}: {high_description}
Scoring guidance: {scoring_guidance}

--- CALIBRATION ANCHORS ---
Example scoring low (~ 1): {low_example}
Example scoring high (~ 4): {high_example}

--- SCORING RULES ---
Use integer scores only: 0, 1, 2, 3, 4, 5.
- 0-1: very little or none of this quality
- 2-3: moderate or mixed presence
- 4-5: strongly exhibits this quality
Reserve 0 and 5 for strong, unambiguous cases.

--- OPTION TO SCORE ---
{option_text}

--- RESPONSE FORMAT ---
Return valid JSON only (no markdown fences, no extra keys):
{{
  "dimension_id": {dimension_id},
  "score": <0 | 1 | 2 | 3 | 4 | 5>,
  "justification": "<one short sentence with evidence from the option>",
  "zero_reason": "<midpoint | not_applicable | uncertain | null>"
}}

Rules for zero_reason:
- Use "midpoint" when score is 2 or 3 due to mixed signals.
- Use "not_applicable" if the dimension does not meaningfully apply.
- Use "uncertain" if evidence is insufficient.
- Otherwise use null.
"""
\end{lstlisting}

\subsubsection*{Alternative pipeline: direct dimension generation.}
The \texttt{method\_llm\_gen/} pipeline also supports generating dimensions directly from a domain description, without first eliciting and condensing reasons. We do not use this path for the deployed studies but include the prompt for completeness.

\noindent\texttt{method\_llm\_gen/prompts/llm\_pref\_gen.txt}: direct dimension-generation prompt.
\begin{lstlisting}
You are helping design a preference elicitation study for rapid personalization.

DOMAIN: {domain}
CHOICE CONTEXT: {choice_context}
NUMBER OF DIMENSIONS: {N}

Core requirement: each dimension must encode a REAL TRADEOFF.
For every dimension, both poles must be actively preferred by real people:
- LOW-pole fans give up something the HIGH pole offers
- HIGH-pole fans give up something the LOW pole offers

Reject dimensions where one side is just objectively better (quality axes),
for example "high quality vs low quality" or any near-universal preference.

Design criteria:
- Directional disagreement: people prefer opposite directions, not only different intensity
- Consequential: helps predict pairwise choices
- Specific: can be scored consistently from option text
- Distinct: minimize redundancy across dimensions

First write 2-4 sentences of reasoning about the key tradeoffs in this domain.
Then produce exactly {N} dimensions.
Then include a redundancy check with any correlated pairs and a disentanglement suggestion.

Respond with valid JSON only (no markdown, no extra prose).

{
  "domain": string,
  "choice_context": string,
  "reasoning": string,
  "dimensions": [
    {
      "id": integer,
      "name": string,
      "low_pole": {
        "label": string,          // 2-5 word name for this end of the scale
        "description": string,    // what an option at this pole looks like
        "typical_person": string  // who tends to prefer this end
      },
      "high_pole": {
        "label": string,
        "description": string,
        "typical_person": string
      },
      "example_contrast": {
        "low_option": string,     // a concrete example option at the low pole
        "high_option": string     // a concrete example option at the high pole
      },
      "articulability": "explicit" | "partially_explicit" | "implicit",
      "estimated_variance": "mostly_between_person" | "mostly_between_option" | "both",
      "scoring_guidance": string  // 1-2 sentences describing what distinguishes
                                  // a 1 from a 4 from a 7 on this dimension,
                                  // without mentioning scores explicitly
    }
  ],
  "redundancy_check": [
    {
      "dimension_ids": [integer, integer],
      "correlation_note": string,
      "disentanglement_suggestion": string
    }
  ]
}
\end{lstlisting}

\subsection*{Per-domain basis evaluation}

We benchmark each domain's discovered basis against the variance-maximizing rank-$K$ subspace of the choice covariance produced by principal components analysis (PCA). Two diagnostics are useful: a \emph{scree plot} of the choice-covariance eigenvalue spectrum (which tells us how concentrated true-choice variance is in the encoder feature space, independent of the discovered basis) and a \emph{per-dimension breakdown} that shows, for each named dimension, how much choice variance it captures, how independent it is of the rest of the basis, and (for the two online domains) the cosine between the ridge-fit direction and a contrastive ``mean of high-scoring options minus mean of low-scoring options'' check. Together these diagnostics let us read off three things: (i) whether the encoder's choice-covariance has heavy enough head structure for a low-rank basis to make sense; (ii) whether each named dimension is doing useful, non-redundant work in the basis; and (iii) whether the ridge fit recovered the same direction the corpus extremes alone would suggest.

\begin{figure}[H]
  \centering
  \includegraphics[width=0.98\linewidth]{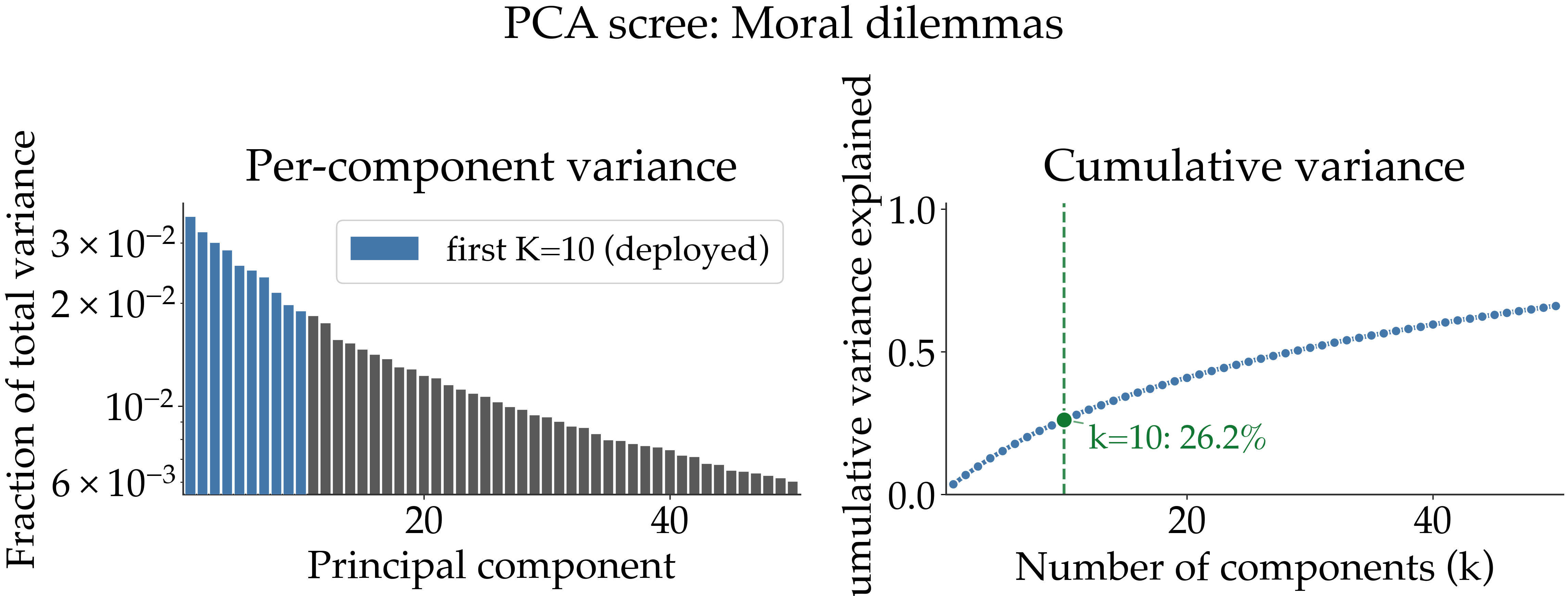}
  \caption{PCA scree plot for moral dilemmas ($N=300$ corpus actions, $d=4096$). Left: per-component variance fraction (log $y$-axis). Right: cumulative variance with thresholds at 50/75/90/95/99\%.}
  \label{fig:scree-dailydilemmas}
\end{figure}

\begin{figure}[H]
  \centering
  \includegraphics[width=0.98\linewidth]{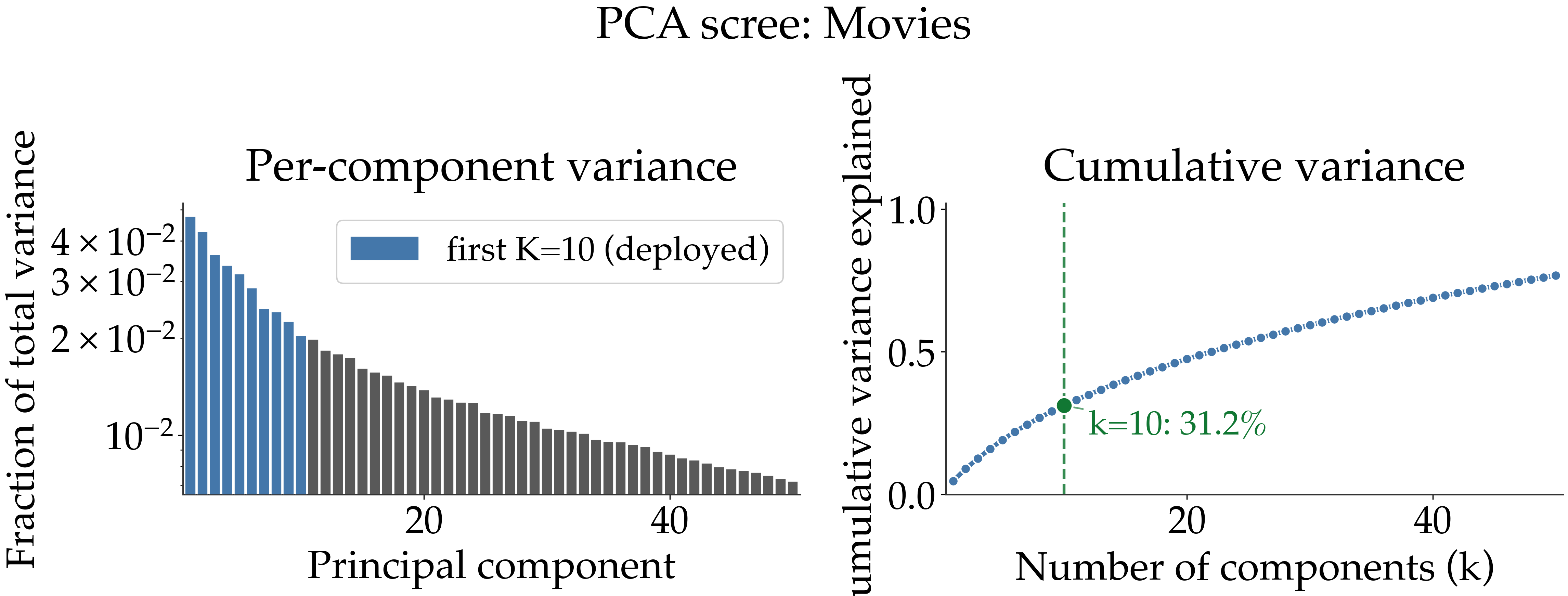}
  \caption{PCA scree plot for movies ($N=100$ films, $d=4096$). Same conventions as Figure~\ref{fig:scree-dailydilemmas}.}
  \label{fig:scree-movies-100}
\end{figure}

\begin{figure}[H]
  \centering
  \includegraphics[width=0.98\linewidth]{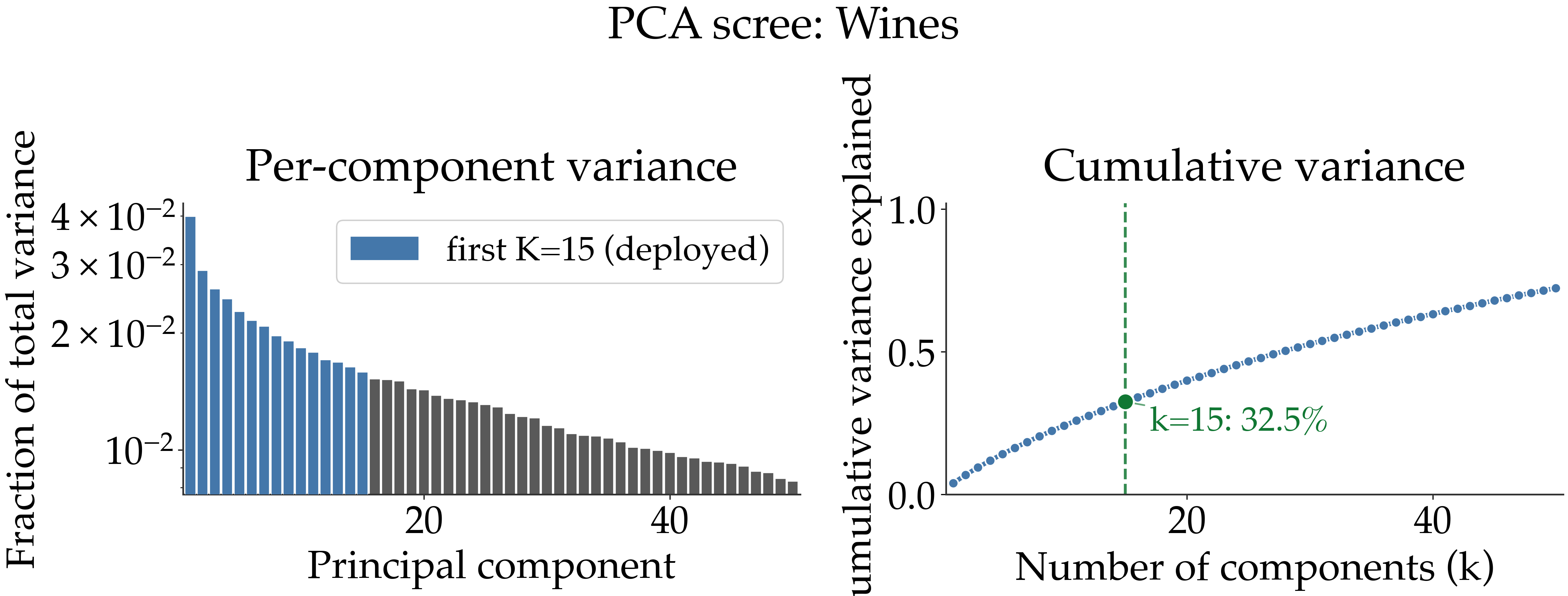}
  \caption{PCA scree plot for wines ($N=100$ wines, $d=4096$). Same conventions as Figure~\ref{fig:scree-dailydilemmas}.}
  \label{fig:scree-wines-100}
\end{figure}

\begin{figure}[H]
  \centering
  \includegraphics[width=0.98\linewidth]{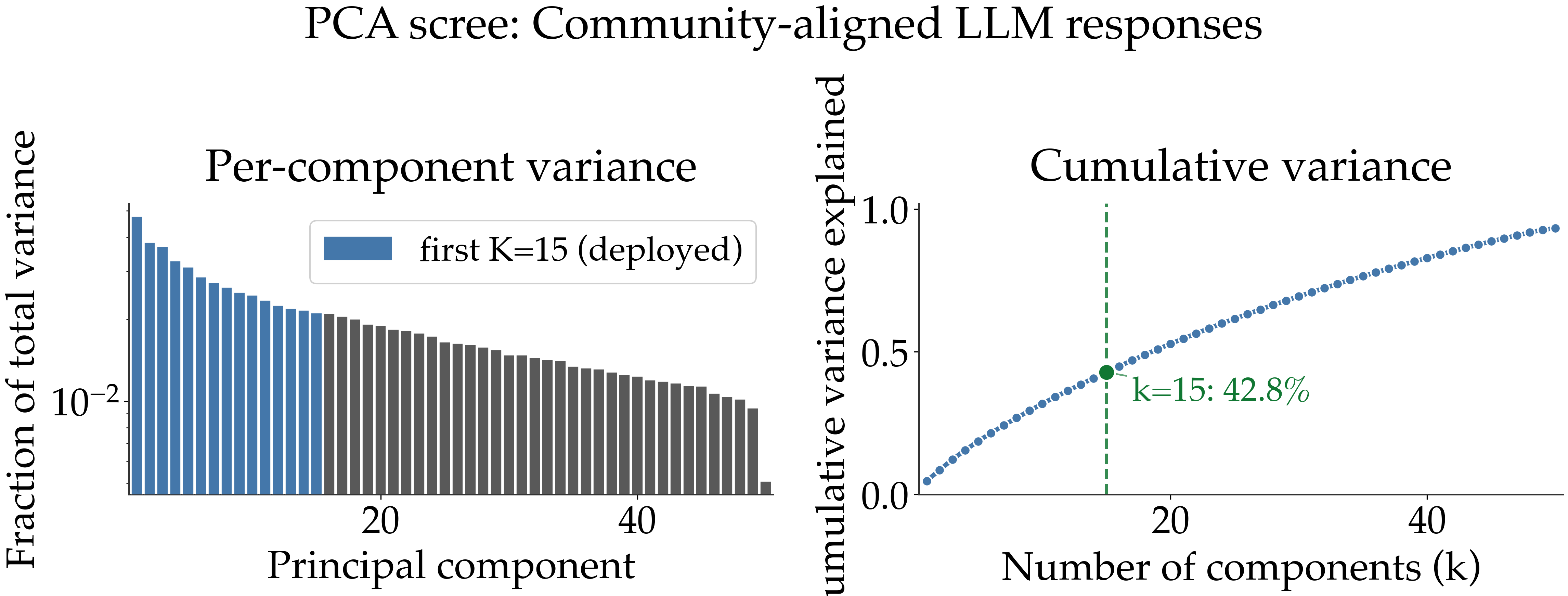}
  \caption{PCA scree plot for LLM responses ($N=50$ responses, $d=4096$). Same conventions as Figure~\ref{fig:scree-dailydilemmas}.}
  \label{fig:scree-coalign-50}
\end{figure}

\paragraph{Scree plots.} Figures~\ref{fig:scree-dailydilemmas}--\ref{fig:scree-coalign-50} show the PCA spectrum for each domain. The left panel of each figure plots the variance fraction captured by each principal component (log scale); the right panel shows cumulative variance with thresholds annotated. All four domains exhibit a smooth, slowly decaying spectrum without a sharp elbow, which is consistent with our observation in Section~\ref{sec:decomp} that no rank-10 or rank-15 basis (LLM-discovered or PCA-derived) captures a majority of the choice covariance, and that the value of the discovered basis lies in interpretability and downstream sample efficiency rather than maximal variance coverage.

\paragraph{Per-dimension breakdown.} Tables~\ref{tab:perdim-dailydilemmas}--\ref{tab:perdim-coalign-50} report, for each named dimension in each domain, its in-corpus variance share, its per-dimension independence, and the cumulative ratio $r_j$ against the rank-$j$ PCA optimum. For the two online domains we additionally report the cosine between the ridge-fit direction $v_k$ and a contrastive mean-difference vector (mean embedding of the top-scoring options minus mean of the bottom-scoring options on dimension $k$). Cosines mostly fall in the $0.5$--$0.85$ range, indicating that the ridge fit recovers a direction that aligns substantially with the contrast the corpus's extremes alone would suggest, while differing from it where the ridge fit benefits from leveraging information from intermediate-scoring options.

\begin{table}[H]
  \centering
  \footnotesize
  \caption{Per-dimension basis evaluation for Moral dilemmas (the deployed $K{=}10$ basis). Rows are ordered by in-corpus projection variance. \textit{var.} is the dimension's share of total choice variance; \textit{indep.} is per-dimension independence (1 = uncorrelated with the rest of $V$, 0 = redundant); \textit{cum.\ $r_j$} is the cumulative rank-$j$ ratio against the PCA optimum (100\% = full overlap with the top-$j$ PCA subspace). \textit{cos.} is the cosine between the ridge-fit direction $v_k$ and a contrastive mean-difference vector (mean of the top-scoring options minus mean of the bottom-scoring), as a sanity check that ridge fitting recovers the same direction the corpus extremes suggest.}
  \label{tab:perdim-dailydilemmas}
  \begin{tabular*}{\linewidth}{@{\extracolsep{\fill}}r l r r r r@{}}
    \toprule
    rank & dimension & var. & indep. & cum.\ $r_j$ & cos. \\
    \midrule
    1 & Privacy & 2.23\% & 0.873 & 57.6\% & 0.80 \\
    2 & Informality & 1.88\% & 0.855 & 55.8\% & 0.83 \\
    3 & Tradition Adherence & 1.86\% & 0.900 & 56.2\% & 0.83 \\
    4 & Financial Prudence & 1.66\% & 0.923 & 55.6\% & 0.79 \\
    5 & Efficiency & 1.58\% & 0.756 & 55.8\% & 0.72 \\
    6 & Decisiveness & 1.57\% & 0.882 & 56.1\% & 0.75 \\
    7 & Moral Integrity & 1.56\% & 0.960 & 56.6\% & 0.78 \\
    8 & Creativity & 1.50\% & 0.872 & 57.4\% & 0.77 \\
    9 & Social Harmony & 1.27\% & 0.850 & 57.5\% & 0.76 \\
    10 & Structure Preference & 1.26\% & 0.889 & 57.8\% & 0.76 \\
    \bottomrule
  \end{tabular*}
\end{table}

\begin{table}[H]
  \centering
  \footnotesize
  \caption{Per-dimension basis evaluation for Movies (the deployed $K{=}10$ basis). Rows are ordered by in-corpus projection variance. \textit{var.} is the dimension's share of total choice variance; \textit{indep.} is per-dimension independence (1 = uncorrelated with the rest of $V$, 0 = redundant); \textit{cum.\ $r_j$} is the cumulative rank-$j$ ratio against the PCA optimum (100\% = full overlap with the top-$j$ PCA subspace). \textit{cos.} is the cosine between the ridge-fit direction $v_k$ and a contrastive mean-difference vector (mean of the top-scoring options minus mean of the bottom-scoring), as a sanity check that ridge fitting recovers the same direction the corpus extremes suggest.}
  \label{tab:perdim-movies-100}
  \begin{tabular*}{\linewidth}{@{\extracolsep{\fill}}r l r r r r@{}}
    \toprule
    rank & dimension & var. & indep. & cum.\ $r_j$ & cos. \\
    \midrule
    1 & Humor Intensity & 2.18\% & 0.739 & 43.0\% & 0.68 \\
    2 & Action Intensity & 1.91\% & 0.796 & 42.6\% & 0.62 \\
    3 & Emotional Depth & 1.89\% & 0.744 & 44.5\% & 0.57 \\
    4 & Suspense/Atmosphere & 1.84\% & 0.779 & 46.0\% & 0.70 \\
    5 & Sci-Fi/Fantasy Worldbuilding & 1.55\% & 0.839 & 46.0\% & 0.56 \\
    6 & Family-Friendly Content & 1.32\% & 0.899 & 45.7\% & 0.65 \\
    7 & Visual Spectacle & 1.26\% & 0.822 & 45.9\% & 0.63 \\
    8 & Historical Authenticity & 1.25\% & 0.848 & 46.2\% & 0.57 \\
    9 & Adventure Scope & 1.23\% & 0.893 & 46.6\% & 0.59 \\
    10 & Survival/Stress Scenarios & 1.06\% & 0.986 & 46.8\% & 0.57 \\
    \bottomrule
  \end{tabular*}
\end{table}

\begin{table}[H]
  \centering
  \footnotesize
  \caption{Per-dimension basis evaluation for Wines (the deployed $K{=}15$ basis). Rows are ordered by in-corpus projection variance. \textit{var.} is the dimension's share of total choice variance; \textit{indep.} is per-dimension independence (1 = uncorrelated with the rest of $V$, 0 = redundant); \textit{cum.\ $r_j$} is the cumulative rank-$j$ ratio against the PCA optimum (100\% = full overlap with the top-$j$ PCA subspace).}
  \label{tab:perdim-wines-100}
  \begin{tabular*}{\linewidth}{@{\extracolsep{\fill}}r l r r r@{}}
    \toprule
    rank & dimension & var. & indep. & cum.\ $r_j$ \\
    \midrule
    1 & Traditional Style & 1.50\% & 0.851 & 36.9\% \\
    2 & Sweetness & 1.49\% & 0.886 & 42.7\% \\
    3 & Fruit Intensity & 1.39\% & 0.887 & 45.4\% \\
    4 & Oak Influence & 1.36\% & 0.943 & 47.4\% \\
    5 & Body and Structure & 1.31\% & 0.822 & 48.9\% \\
    6 & Floral Aroma Intensity & 1.20\% & 0.898 & 49.7\% \\
    7 & Tropical Aroma Intensity & 1.16\% & 0.850 & 50.3\% \\
    8 & Value for Money & 1.10\% & 0.958 & 50.7\% \\
    9 & Playfulness & 1.09\% & 0.951 & 51.2\% \\
    10 & Earthy/Savory Character & 1.07\% & 0.849 & 51.7\% \\
    11 & Effervescence & 1.05\% & 0.941 & 52.2\% \\
    12 & Acidity & 1.00\% & 0.898 & 52.5\% \\
    13 & Aromatic Complexity & 0.94\% & 0.950 & 52.7\% \\
    14 & Aging Potential & 0.94\% & 0.940 & 52.9\% \\
    15 & Sessionability & 0.91\% & 0.912 & 53.0\% \\
    \bottomrule
  \end{tabular*}
\end{table}

\begin{table}[H]
  \centering
  \footnotesize
  \caption{Per-dimension basis evaluation for LLM responses (the deployed $K{=}15$ basis). Rows are ordered by in-corpus projection variance. \textit{var.} is the dimension's share of total choice variance; \textit{indep.} is per-dimension independence (1 = uncorrelated with the rest of $V$, 0 = redundant); \textit{cum.\ $r_j$} is the cumulative rank-$j$ ratio against the PCA optimum (100\% = full overlap with the top-$j$ PCA subspace).}
  \label{tab:perdim-coalign-50}
  \begin{tabular*}{\linewidth}{@{\extracolsep{\fill}}r l r r r@{}}
    \toprule
    rank & dimension & var. & indep. & cum.\ $r_j$ \\
    \midrule
    1 & Practicality & 2.48\% & 0.380 & 29.0\% \\
    2 & Authenticity & 1.84\% & 0.404 & 28.0\% \\
    3 & Community Focus & 1.48\% & 0.536 & 26.3\% \\
    4 & Structure & 0.85\% & 0.621 & 23.8\% \\
    5 & Clarity & 0.78\% & 0.498 & 22.2\% \\
    6 & Emotional Resonance & 0.52\% & 0.402 & 20.6\% \\
    7 & Sustainability & 0.52\% & 0.647 & 19.4\% \\
    8 & Historical Context & 0.51\% & 0.554 & 18.6\% \\
    9 & Actionability & 0.47\% & 0.457 & 17.9\% \\
    10 & Conciseness & 0.45\% & 0.470 & 17.3\% \\
    11 & Formality & 0.37\% & 0.503 & 16.7\% \\
    12 & Depth & 0.26\% & 0.660 & 16.1\% \\
    13 & Descriptiveness & 0.23\% & 0.668 & 15.5\% \\
    14 & Efficiency & 0.20\% & 0.609 & 15.0\% \\
    15 & Creativity & 0.16\% & 0.688 & 14.4\% \\
    \bottomrule
  \end{tabular*}
\end{table}

\section{Experimental interface}
\label{app:interface}

This appendix shows screenshots of the experimental interface that participants saw during the online studies. The same procedure ran in both domains; movies and dilemmas screenshots are interleaved here to show how the interface adapts the per-domain stimuli without changing the underlying flow. The trial-rendering code is in \texttt{web-interface/index.html}; the per-trial logic and category-midpoint construction are in the Python and JavaScript implementations described in Appendix~\ref{app:implementation}.

\paragraph{Practice phase.} Each session begins with $T_{\text{prac}}=5$ practice trials, after which participants proceed to the main task. Practice trials show one dimension at a time and ask which option is more aligned with that dimension's high pole; correctness feedback is immediate. The intent is to familiarize participants with the named dimensions before they start the main task.

\begin{figure}[H]
  \centering
  \includegraphics[width=0.75\linewidth]{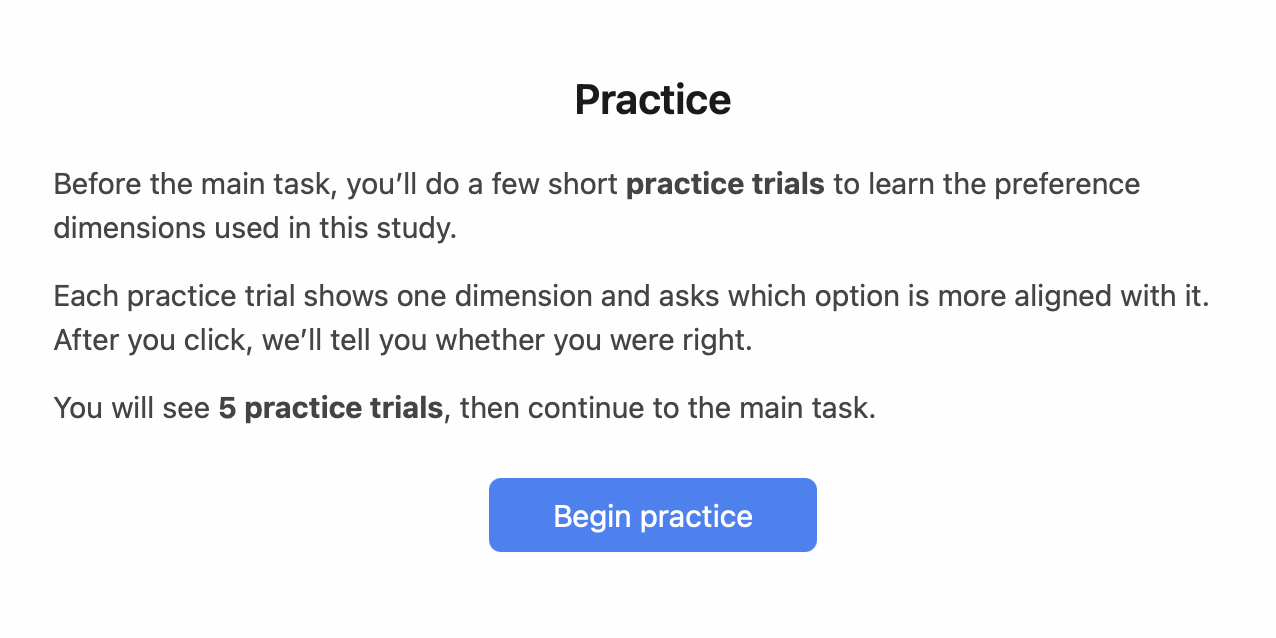}
  \caption{Practice-phase introduction screen, identical across all conditions. Participants are told they will see five practice trials before the main task, and that each trial shows one dimension and asks which option is more aligned with it.}
  \label{fig:qt-practice-intro}
\end{figure}

\begin{figure}[H]
  \centering
  \includegraphics[width=0.65\linewidth]{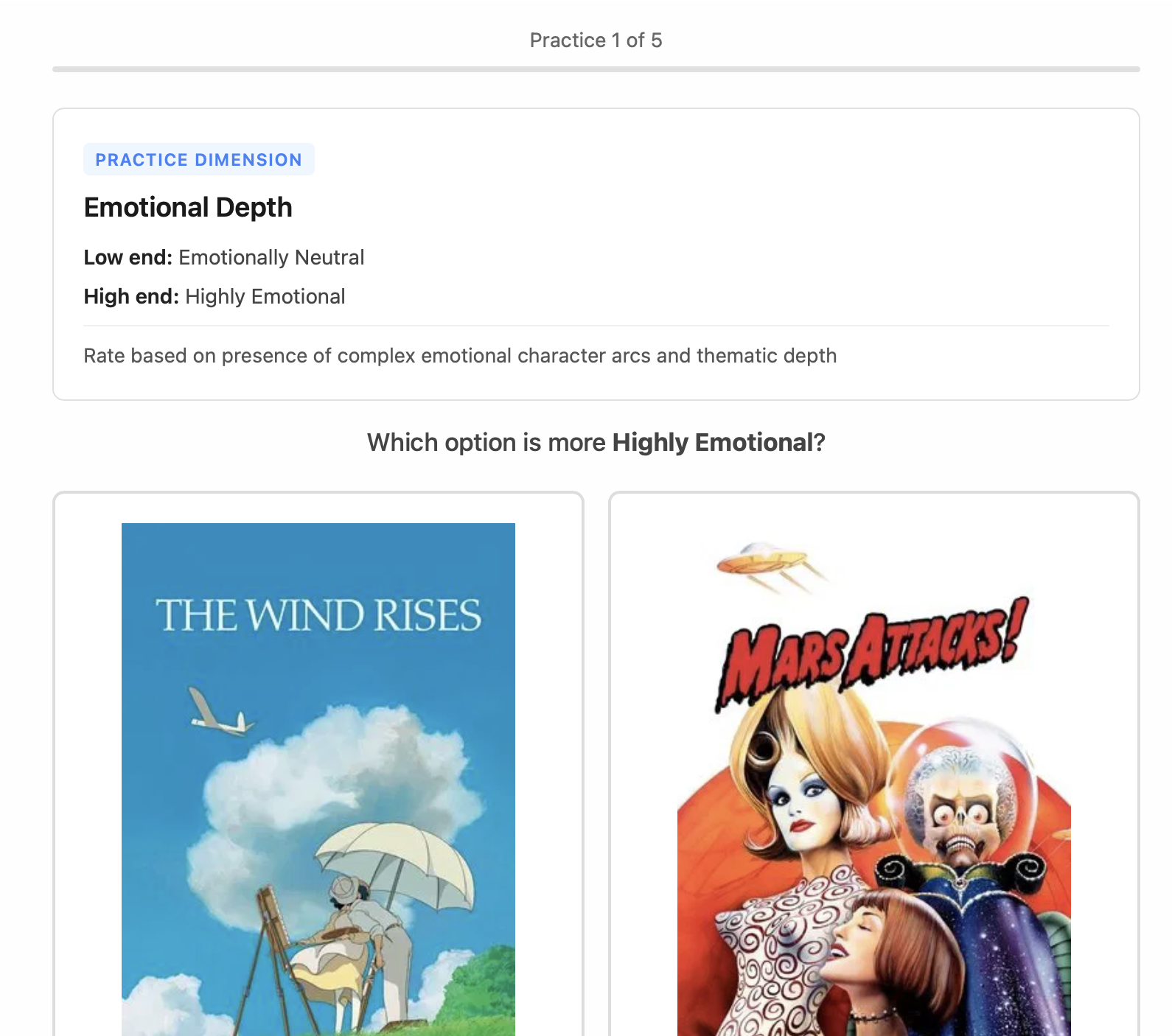}
  \caption{A practice trial in the movies domain (trial 1 of 5). The visible dimension (\emph{Emotional Depth}, with low/high anchor labels and the LLM-generated scoring guidance) is shown in a card above the two options. The participant clicks the option more aligned with the high pole; feedback (correct/incorrect, with the ground-truth dimension score) appears after the click.}
  \label{fig:qt-practice-trial}
\end{figure}

\paragraph{Main task.} After practice, participants proceed to $T=20$ binary-choice trials. The intro screen describes how the inference UI will appear after each choice (in the inference conditions) or notes that no per-trial feedback will be requested (in \textsc{choice\_only}).

\begin{figure}[H]
  \centering
  \includegraphics[width=0.75\linewidth]{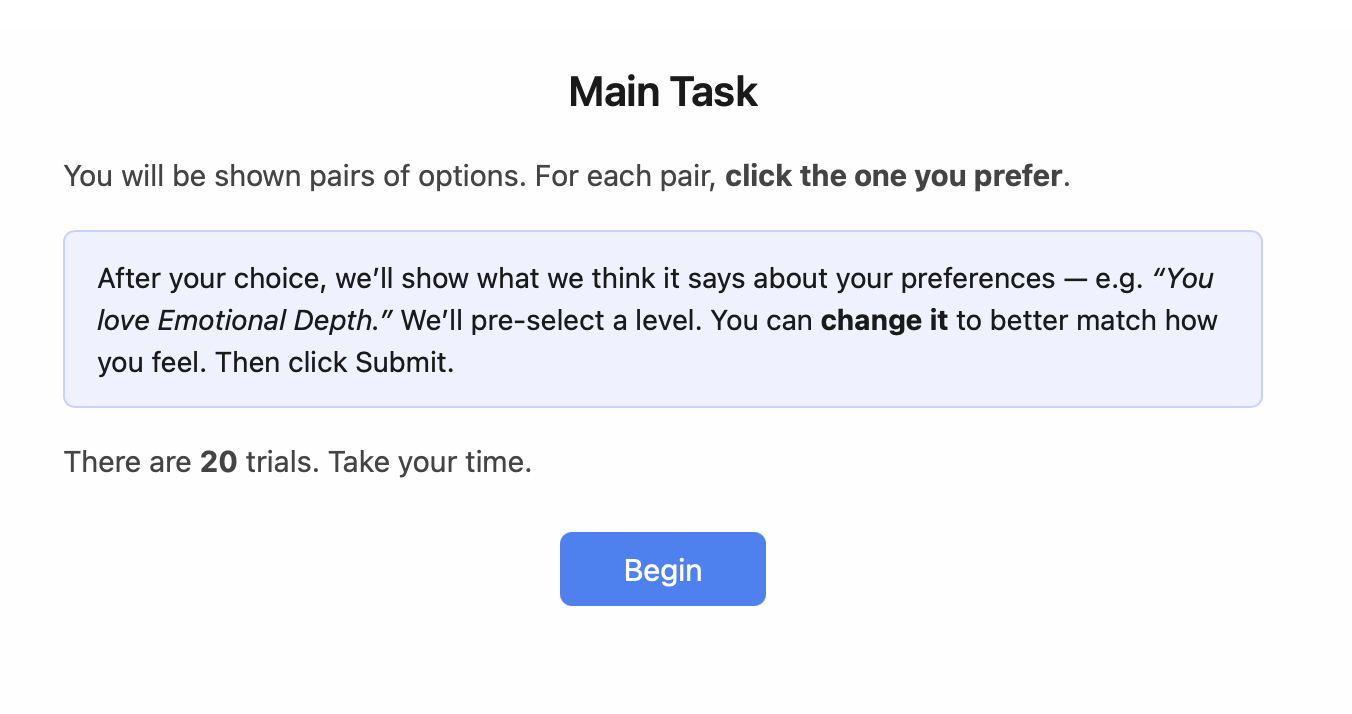}
  \caption{Main-task introduction screen for an inference condition. Participants are told to click the option they prefer, and that the model will then show what it thinks the choice says about their preferences and pre-select a level the participant can adjust before submitting.}
  \label{fig:qt-main-intro}
\end{figure}

\begin{figure}[H]
  \centering
  \includegraphics[width=0.6\linewidth]{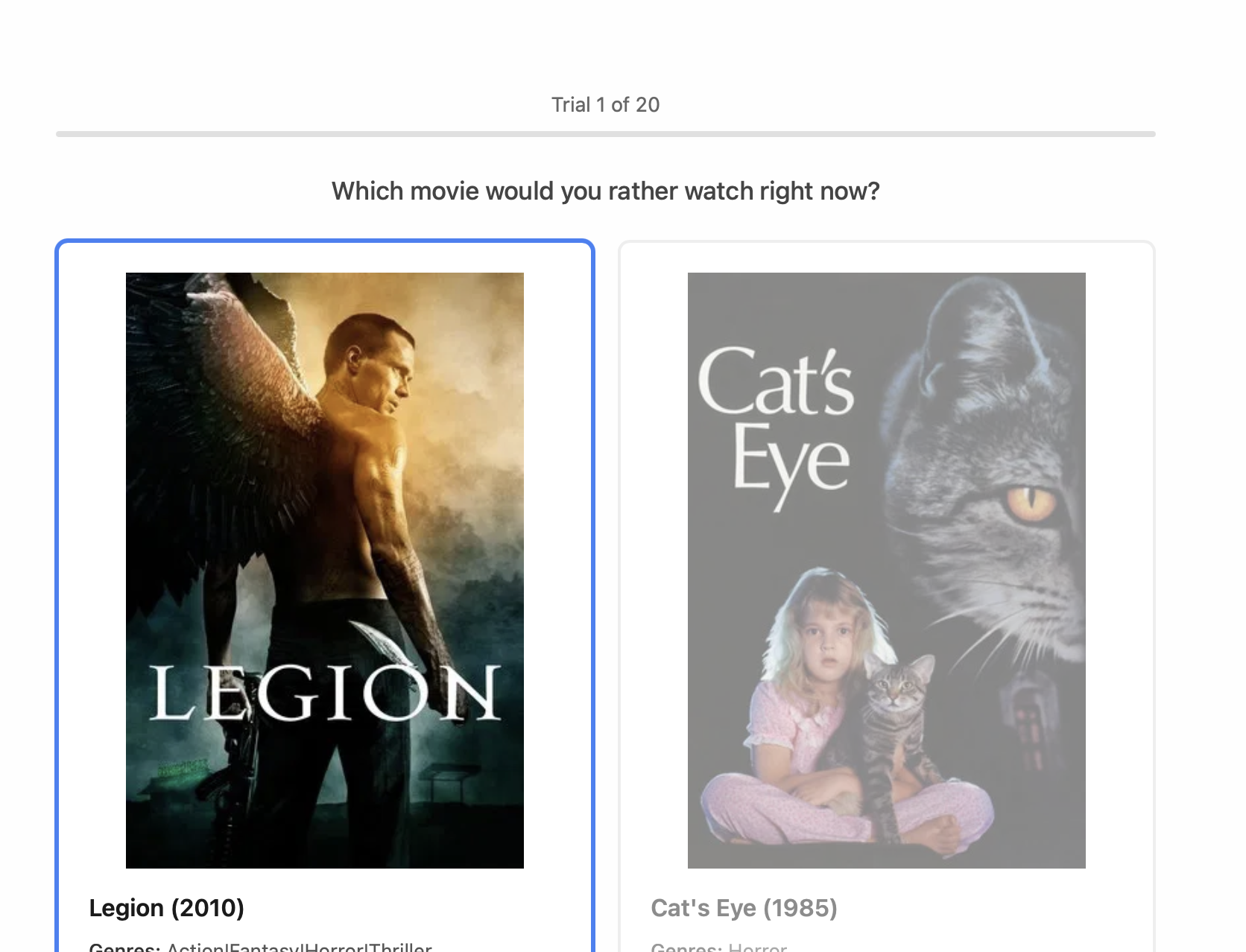}
  \caption{A main-task binary-choice trial (movies, trial 1 of 20). The participant clicks the option they would rather watch; the chosen option is highlighted in blue. In the \textsc{choice\_only} condition the trial advances after the click; in inference conditions the chosen option triggers the feedback panel shown next.}
  \label{fig:qt-main-trial}
\end{figure}

\paragraph{Feedback elicitation (inference conditions).} In \textsc{inference\_categories} and \textsc{inference\_affirm}, after the binary choice the participant sees a feedback panel listing the top-3 dimensions for the trial (ranked by the per-dimension--normalized projection $|u_{t,k}|/m_k$) along with a model-pre-selected level for each. Participants affirm, adjust, or remove the pre-selected inferences before submitting (affirm/remove in \textsc{inference\_affirm}, adjust the level in \textsc{inference\_categories}).

\begin{figure}[H]
  \centering
  \includegraphics[width=0.6\linewidth]{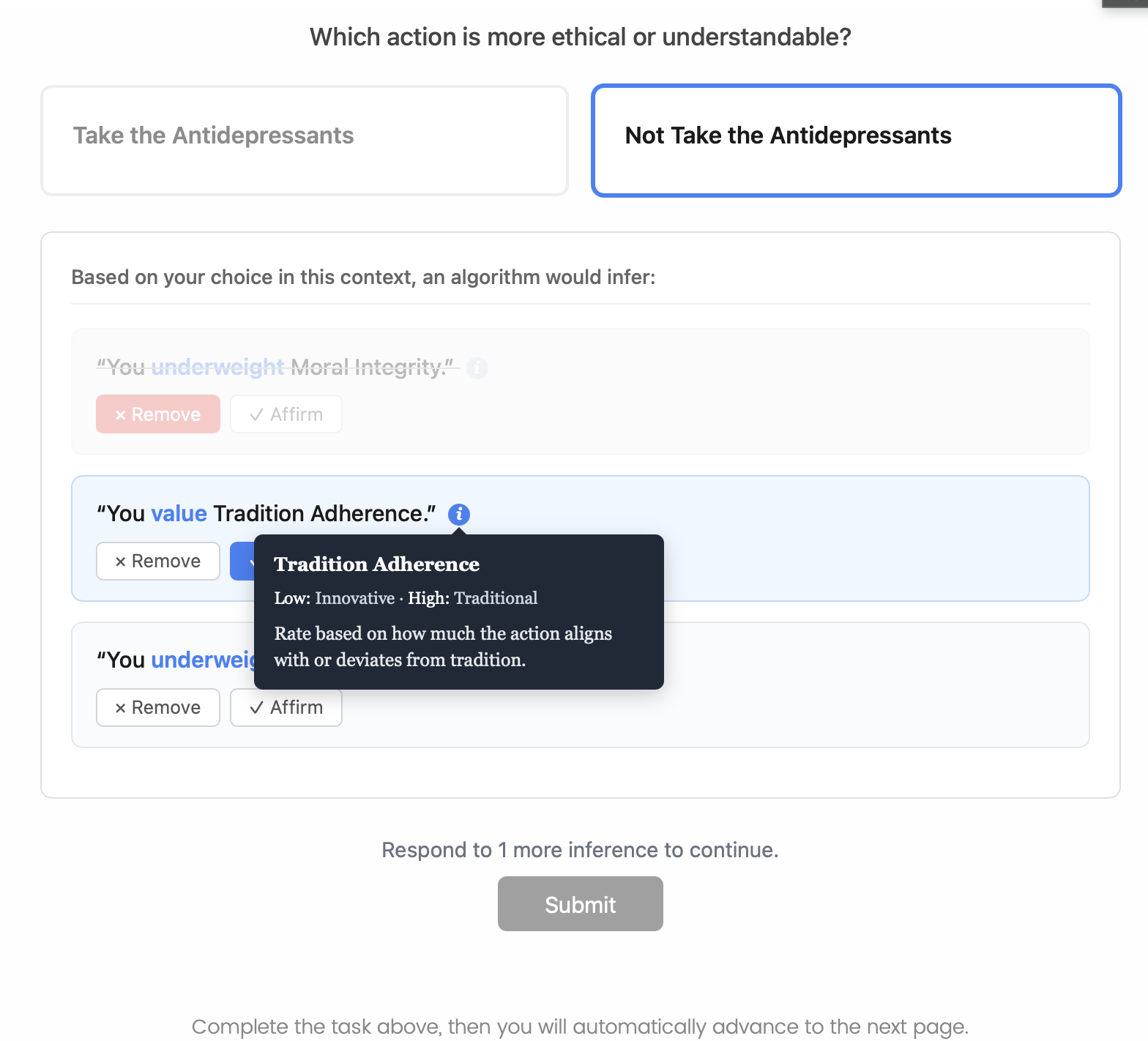}
  \caption{Feedback panel in the \textsc{inference\_affirm} condition (moral dilemmas). The model proposes a natural-language inference for each visible dimension; the format is \emph{``You [value level] [DimensionName]''} (e.g., ``You value Tradition Adherence''). Participants can affirm the inference or remove it. Hovering on the info icon (shown for \emph{Tradition Adherence}) reveals the dimension's full definition.}
  \label{fig:qt-affirm}
\end{figure}

\paragraph{Post-task summary comparison and held-out predictions.} After all $T$ trials, participants see two ``post-experiment'' screens that constitute the H2 and H3 measures. First, they compare two natural-language summaries of their preferences (one from the augmented model, one from the baseline; left/right randomized). Second, they rate the accuracy of two held-out predictions---one from each model---on pairs selected so the two models disagree, and one is maximally confident.

\begin{figure}[H]
  \centering
  \includegraphics[width=0.65\linewidth]{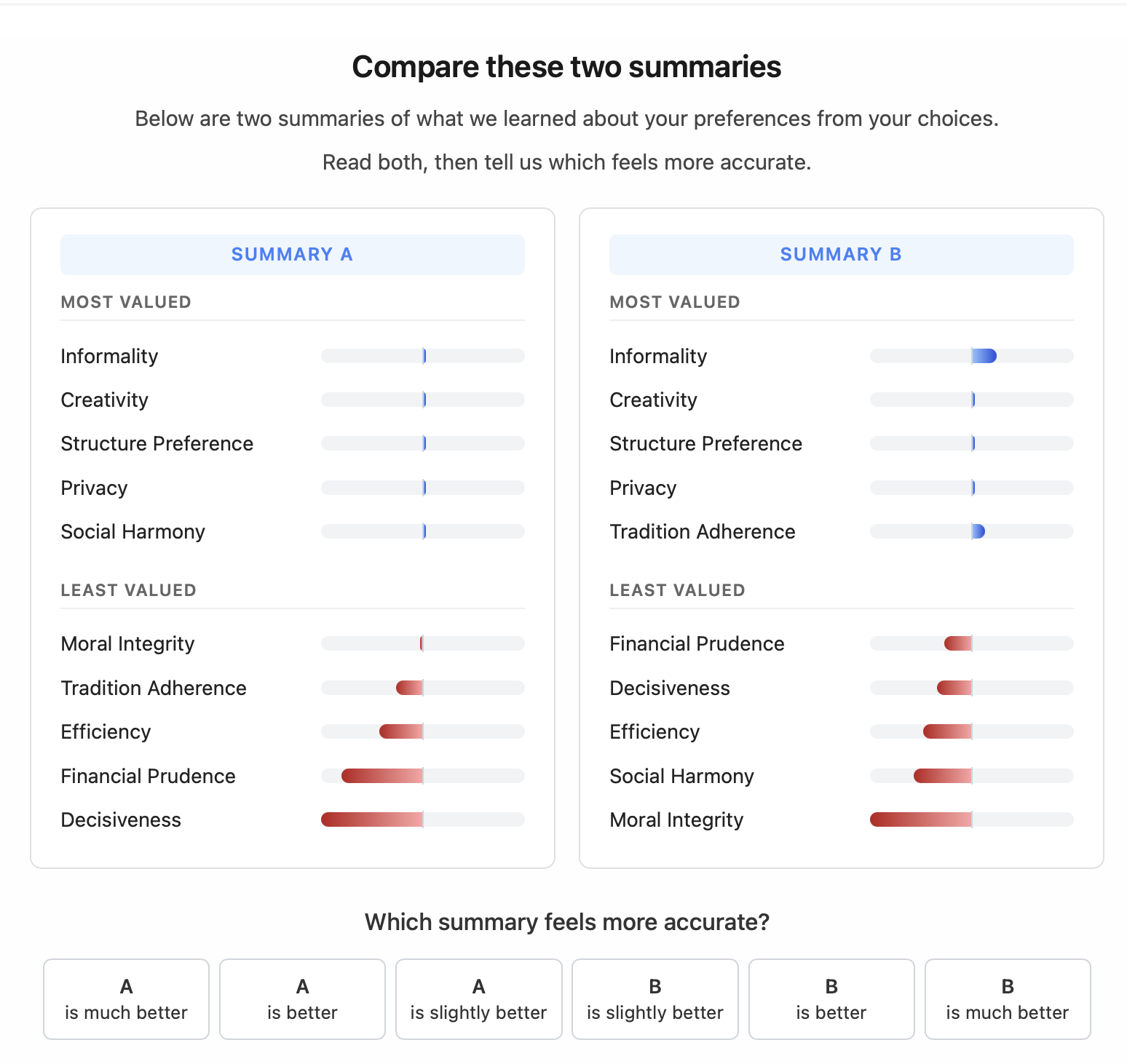}
  \caption{Summary-comparison screen (moral dilemmas; H2 dependent measure). Two summaries display each participant's most- and least-valued dimensions, rendered as horizontal bar charts. The participant rates which summary feels more accurate on a signed 6-point scale (A is much better $\to$ B is much better). Augmented and baseline assignment to A/B is randomized per participant.}
  \label{fig:qt-h2}
\end{figure}

\begin{figure}[H]
  \centering
  \includegraphics[width=0.65\linewidth]{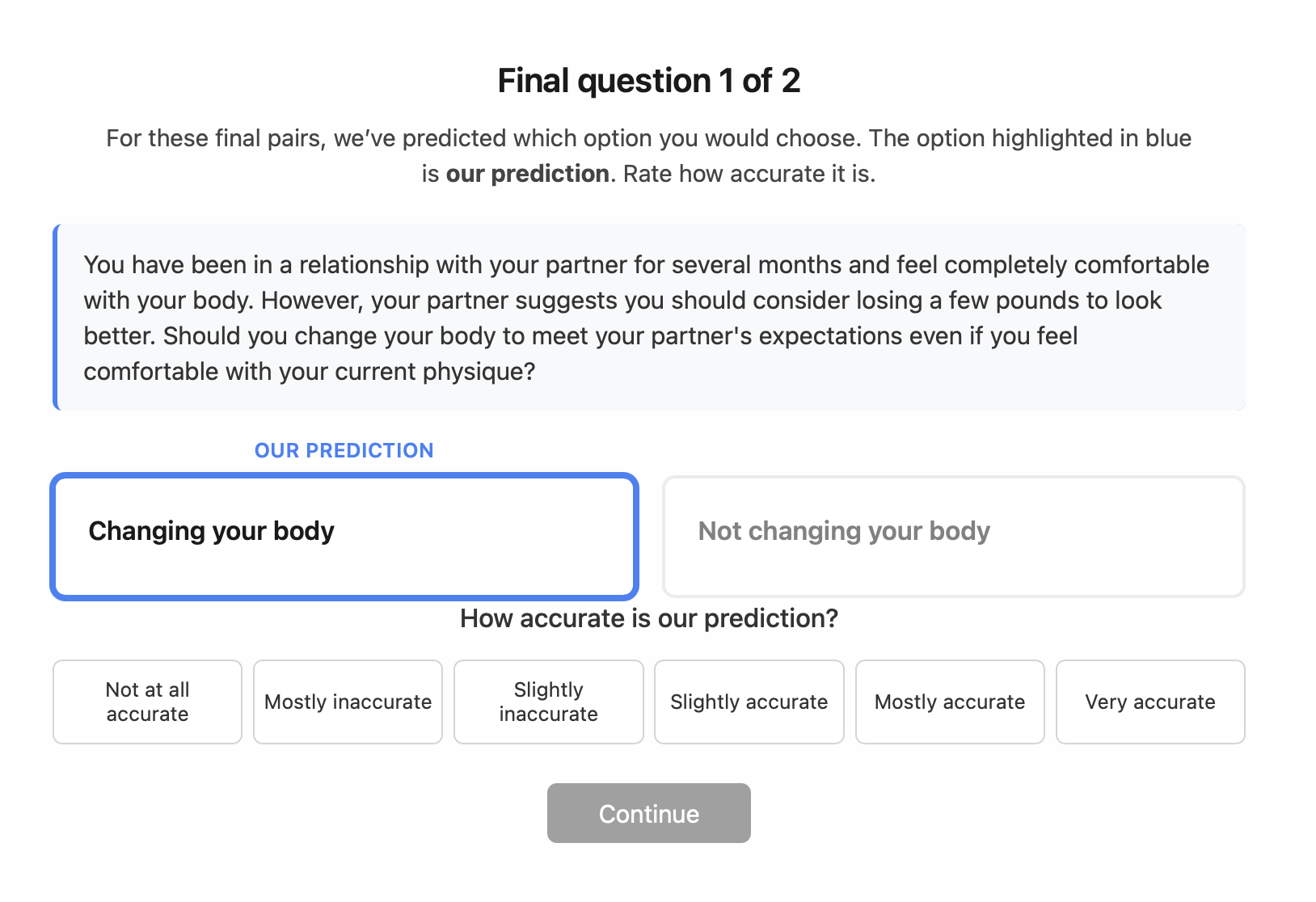}
  \caption{Held-out prediction screen (moral dilemmas; H3 dependent measure). For two pairs from a held-out pool on which the augmented and baseline models disagree, the participant sees one model's predicted choice (highlighted in blue) and rates ``How accurate is our prediction?'' on a 6-point scale (Not at all accurate $\to$ Very accurate). The order in which the augmented vs.\ baseline prediction is rated is randomized across participants so that model and presentation order are crossed.}
  \label{fig:qt-h3}
\end{figure}


\end{document}